\documentclass{article}

\usepackage{arxiv}
\usepackage[utf8]{inputenc} 
\usepackage[T1]{fontenc}    
\usepackage{hyperref}       
\usepackage{url}            
\usepackage{booktabs}       
\usepackage{amsfonts}       
\usepackage{nicefrac}       
\usepackage{microtype}      
\usepackage{lipsum}		
\usepackage{graphicx}
\usepackage{natbib}
\usepackage{doi}
\usepackage{tikz}
\usepackage{longtable}

\title{LLMs can realize combinatorial creativity: generating creative ideas via LLMs for scientific research}

\author{ {\hspace{1mm}Tianyang Gu}\thanks{This work was partially conducted during the studies at Boston University.} \\
	Department of Computer Science\\
    Graduate School of Arts and Sciences, Boston University\\
    Boston, MA\\
	\texttt{tygu@bu.edu} \\
	\And
    {
    \hspace{1mm}Jingjin Wang} \\
	Department of Computer Science\\
	University of Illinois at Urbana-Champaign\\
	  Champaign, IL 61820 \\
	\texttt{jingjin9@illinois.edu} \\
    \And
      {
    \hspace{1mm}Zhihao Zhang} \\
	Department of Electrical and Computer Engineering\\
	University of Delaware\\
	Newark, DE \\
	\texttt{zzhsaga@udel.edu} \\
    \And
      {
    \hspace{1mm}HaoHong Li} \\
	Acaciawood Preparatory Academy\\
	Anaheim, California\\
	\texttt{sherlockl@awschool.org} \\
}

\date{}

\renewcommand{\shorttitle}{}
\hypersetup{
pdftitle={A template for the arxiv style},
pdfsubject={q-bio.NC, q-bio.QM},
pdfauthor={David S.~Hippocampus, Elias D.~Striatum},
pdfkeywords={First keyword, Second keyword, More},
}

\begin{document}
\maketitle

\begin{abstract}
Scientific idea generation has been extensively studied in creativity theory and computational creativity research, providing valuable frameworks for understanding and implementing creative processes. However, recent work using Large Language Models (LLMs) for research idea generation often overlooks these theoretical foundations. We present a framework that explicitly implements combinatorial creativity theory using LLMs, featuring a generalization-level retrieval system for cross-domain knowledge discovery and a structured combinatorial process for idea generation. The retrieval system maps concepts across different abstraction levels to enable meaningful connections between disparate domains, while the combinatorial process systematically analyzes and recombines components to generate novel solutions. Experiments on the OAG-Bench dataset demonstrate our framework's effectiveness, consistently outperforming baseline approaches in generating ideas that align with real research developments (improving similarity scores by 7\%-10\% across multiple metrics). Our results provide strong evidence that LLMs can effectively realize combinatorial creativity when guided by appropriate theoretical frameworks, contributing both to practical advancement of AI-assisted research and theoretical understanding of machine creativity.
\end{abstract}


\section{Introduction}
The ability to generate creative ideas is fundamental to human progress, spanning domains from scientific discovery to artistic expression. The emergence of Large Language Models (LLMs) presents unprecedented opportunities for computational systems to engage in creative processes. Recent work has shown promising results in using Large Language Models (LLMs) for creative idea generation (\cite{si2024can, kumar2024can, wang2305scimon}) However, these approaches often lack grounding in established theories of computational creativity, potentially limiting their effectiveness as well as their future development. For example, they only care about the novelty of the generated idea, however, novelty is meaningless if it is valueless. To address this gap, we emphasize the importance of guiding LLMs' ideation process through computational creativity theory, particularly Boden's framework of combinatorial creativity (\cite{boden2004creative}) which describes how novel ideas emerge from combining existing concepts in unexpected ways.

The field of creativity research offers several theoretical frameworks, notably the "four P's" taxonomy: Person, Process, Product, and Press (\cite{rhodes1961analysis}), which was subsequently adapted for computational creativity (\cite{jordanous2016four}). This framework provides multiple perspectives for understanding and evaluating creative systems. Within the process perspective, Boden's theory of conceptual spaces has been influential, distinguishing between three types of creativity: combinatorial, exploratory, and transformational. This paper focuses on combinatorial creativity. The Product perspective argues what product is creative, which is crucial for evaluation. A very common set of creativity criteria are novelty and value.

Building on these established frameworks, we propose an agent-based architecture specifically designed to model and implement combinatorial creativity using LLMs. Our approach systematically maps the theoretical principles of creative cognition onto LLM through prompting, enabling more structured and theoretically grounded idea generation. Also, we developed a retrieval system to get potentially useful concepts for combinatorial creativity. Particularly, the retrieval system can provide concepts from different domains to ensure the creativity process. Through extensive evaluation, we demonstrate that LLMs can effectively realize combinatorial creativity when properly guided by established theoretical frameworks. Our findings contribute both to the practical advancement of LLM-based creative systems and theoretical understanding of machine creativity.

\section{Related work}
\subsection{Computational creativity}
The study of creativity has been systematically organized through the "Four P's" framework (\cite{rhodes1961analysis}), which provides distinct yet interconnected perspectives for understanding creative phenomena: Person, Process, Product, and Press.
\subsubsection{Person}
The Person perspective examines the characteristics of the creative agent, whether human or computational. This view focuses on identifying what traits or capabilities enable creative behavior, which has become increasingly relevant as AI systems take on creative tasks (\cite{jordanous2016four}).
\subsubsection{Process}
The Process perspective encompasses theories about how creative products are generated - specifically examining the cognitive steps and mechanisms involved in creative activities. While many of these theories were originally developed through studying human creativity, they provide valuable frameworks for computational systems.

One of the most influential process theories in computational creativity is Boden's theory of conceptual space, which delineates three types of creativity: combinatorial, exploratory, and transformational. Combinatorial creativity involves connecting familiar ideas in novel ways - a mechanism particularly well-suited to LLMs given their ability to identify and recombine patterns across vast knowledge spaces. Exploratory creativity involves discovering new possibilities within an established conceptual space by testing rule implications, while transformational creativity fundamentally alters the rules to reach previously inaccessible points (\cite{boden2004creative, wiggins2006preliminary}).

The creative process is often conceptualized through iterative frameworks that alternate between generation and evaluation phases. The Geneplore model (\cite{ward1999creative}) involves generating "preinventive structures" through synthesis, transformation, or exemplar retrieval, followed by evaluation and exploration of their properties. Similarly, the Engagement-Reflection (ER) model (\cite{sharpies2013account}) alternates between brainstorming ideas (Engagement) and evaluating their validity (Reflection). These iterative frameworks align well with modern computational approaches that combine generation with critical evaluation.

Stage-based theories provide another perspective on the creative process. Wallas' influential four-stage model describes creativity as progressing through preparation (gathering information), incubation (unconscious processing), inspiration (idea emergence), and verification (development and refinement). While these discrete stages may not map directly to computational systems, they highlight important functional requirements for creative AI: the need to acquire and process relevant knowledge, generate novel connections, and critically evaluate outputs.

More recent work emphasizes the social and interactive nature of creative processes. Glăveanu's perspective-taking theory frames creativity as requiring shifts between creator and audience viewpoints, suggesting the importance of incorporating audience preferences and feedback into computational creative systems. This aligns with modern approaches that use human feedback to guide creative AI systems.

For LLM-based creative systems, these process theories suggest several key design principles:
\begin{itemize}
    \item Structured exploration of conceptual spaces through carefully designed prompting strategies
    \item Implementation of iterative generation-evaluation loops
    \item Incorporation of domain knowledge and audience feedback
    \item Explicit mechanisms for both generating novel combinations and evaluating their value
\end{itemize}

Our work focuses specifically on combinatorial creativity theory, developing methods to help LLMs identify and recombine valuable concepts from scientific literature in theoretically-grounded ways. While existing computational creativity systems often implement iterative generation-evaluation loops following the Geneplore or ER models, our framework takes a direct approach to combinatorial creativity through structured knowledge extraction and guided recombination. This approach provides a focused investigation of how LLMs can implement one specific type of creative process within Boden's theoretical framework.
\subsubsection{Product}
The Product perspective examines what makes outputs worthy of being considered "creative." A dominant framework defines creativity through two fundamental criteria: novelty and value (\cite{boden2004creative, gaut2010philosophy}). Under this definition, a product is considered creative if and only if it satisfies both conditions - being both novel and valuable within its domain.

Novelty itself has multiple important distinctions. Boden differentiates between P-creativity (psychological creativity - novel to the creator) and H-creativity (historical creativity - novel to human history). This basic distinction was further refined by Kaufman and Beghetto's (\cite{kaufman2009beyond}) "Four C's" model, which hierarchically categorizes creativity into Big-C (historically significant achievements), Pro-C (professional-level contributions), Little-c (everyday creative problem-solving), and Mini-c (personal creative insights). (\cite{bartel1985originality}) emphasizes that truly original works must serve as origins - introducing unique attributes that influence subsequent works rather than just being different.

Current work in LLM-based idea generation faces several significant evaluation challenges. Most approaches rely heavily on human evaluation, which can introduce subjective biases and be difficult to standardize across different studies. When automated metrics are used, they typically reduce novelty evaluation to simple semantic similarity measures, failing to capture Boden's deeper notion of meaningful novelty or Bartel's concept of originality as influence. Furthermore, many current approaches focus almost exclusively on novelty while neglecting the crucial value dimension of creativity.

The value criterion, while more challenging to formalize, remains essential for genuine creativity. While some argue that value judgments are inherently subjective and culturally dependent (\cite{moneta1996effect}), others propose that value should be understood in terms of effectiveness or appropriateness to specific goals (\cite{kaufman2012beyond}). However, existing approaches often struggle to systematically assess value, either omitting this criterion entirely or relying solely on subjective human judgments.
For computational creativity systems, particularly those working with LLMs, these theoretical perspectives and current limitations highlight the need for more objective and reproducible evaluation frameworks that can assess both novelty and value. This motivates consideration of new evaluation approaches that could ground assessment in historical scientific development rather than relying purely on similarity metrics or human judgment.
\subsubsection{Press}
The Press perspective considers the broader environmental and cultural context that influences and evaluates creative work. This includes understanding how societal norms, domain conventions, and audience expectations shape both the creative process and the reception of creative products.

This theoretical framework has proven particularly valuable for developing computational creativity systems, as it provides structured ways to think about design choices and evaluation criteria. When applied to LLM-based creative systems, these perspectives help identify key considerations: how to structure the system's capabilities (Person), what processes to implement for idea generation (Process), how to evaluate the quality of generated ideas (Product), and how to ensure the system's outputs are valuable within their intended context (Press).

\subsection{Idea generation via large language models }
Recent advancements in Large Language Models (LLMs) have sparked growing interest in computational idea generation, particularly for scientific research. These efforts can be understood through the lens of computational creativity theory, especially Boden's framework of creative processes. Early work in this direction focused on hypothesis generation (\cite{yang2024leandojo, qi2023large}), which primarily exemplifies exploratory creativity within the confined conceptual space of binary relationships between scientific concepts. While valuable, this approach operates within highly constrained boundaries and doesn't fully leverage LLMs' potential for more transformative creative processes.

More recent approaches have attempted to expand the creative scope by augmenting LLMs with scientific literature. ResearchAgent (\cite{baek2024researchagent}) implements a form of combinatorial creativity by using an entity-centric knowledge store and academic knowledge graphs to help LLMs identify and combine concepts across different papers. This multi-agent framework mirrors the Generation-Evaluation loop described in the Geneplore model, using peer review mechanisms to iteratively refine generated ideas. Similarly, (\cite{wang2305scimon}) develop methods for identifying novel research opportunities by analyzing conceptual relationships across papers, demonstrating how computational systems can systematically explore and combine ideas from existing literature. Comprehensive research automation systems like AI-Scientist (\cite{lu2024ai}) attempt to model the entire creative process, from literature review to idea generation and validation. These approaches align with stage-based creativity theories like Wallas' model, implementing distinct phases for knowledge gathering, idea generation, and verification.

The evaluation of LLM-generated ideas presents unique challenges that intersect with the Product perspective of creativity theory. (\cite{si2024can}) conducted comprehensive comparisons between LLM and human-generated ideas, finding that while LLMs can match human experts in novelty, they often struggle with what creativity theorists term "value" - the practical utility and feasibility of generated ideas. This aligns with broader observations in computational creativity about the challenge of balancing novelty with usefulness.

These existing approaches, while promising, often lack explicit grounding in established creativity theory. Most systems treat idea generation as an information-processing task rather than a creative process that requires systematic exploration and transformation of conceptual spaces. Additionally, current evaluation methods typically focus on practical metrics without considering how these relate to theoretical understandings of creative products. This gap between theory and practice suggests the need for approaches that explicitly incorporate creativity theoretical frameworks into both system design and evaluation.

\begin{figure}[h]
    \centering
    \includegraphics[width=\textwidth]{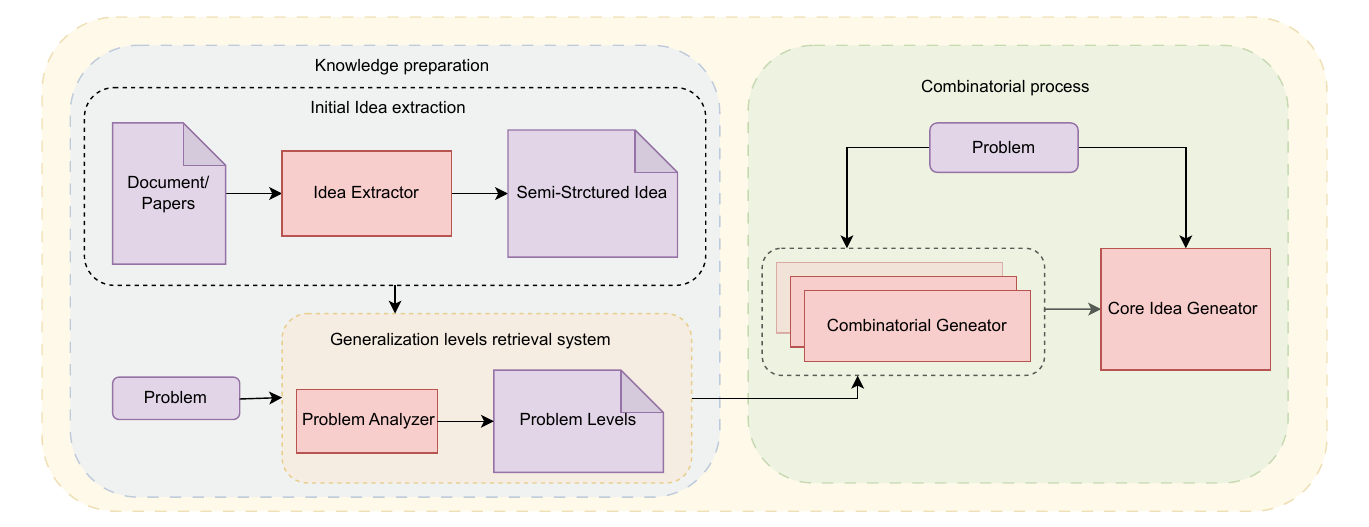} 
    \caption{Combinatorial creativity agent core}
    \label{fig:example}
\end{figure}

\section{Combinatorial creativity agent framework}
Our framework consists of two main phases: knowledge preparation and combinatorial idea generation, as shown in figure 1. The knowledge preparation phase gathers useful familiar concepts, while the combinatorial process uses these concepts to generate creative ideas.

\subsection{Idea Extraction and Retrieval}
Traditional retrieval methods like RAG (Retrieval-Augmented Generation) and academic graph databases often struggle to effectively connect knowledge across different domains, as they typically focus on surface-level similarity or direct keyword matches. To address this challenge, we developed a generalization-level retrieval system, as shown in Figure 2, which enables effective cross-domain knowledge discovery through multi-level semantic matching.

As illustrated in the left part of Figure 2, we introduce a semi-structured "ideation format" that captures both concrete details and abstract principles of innovations. This standardized format stores comprehensive metadata including innovation names, original problems, key mechanisms, novel insights, and most importantly, parallel generalization levels (L1-L4). These levels represent a gradient from domain-specific implementations to universal principles, allowing ideas to be matched at various levels of abstraction. Each innovation entry is structured in a consistent JSON format to ensure reliable processing and comparison.

The retrieval process, shown in the right part of Figure 2 and Figure 1, operates through a two-stage pipeline. First, when a new problem is presented, our AI agent analyzes it using a structured prompt to extract multiple perspectives and problem structures. Each structure is then mapped across the four generalization levels, mirroring the format of stored innovations. In the second stage, we utilize OpenAI's text-embedding-3-large model for high-quality semantic embeddings. The process flows from problem-level embedding through a similarity filter to idea-level embedding. For each problem structure at each generalization level, we compute cosine similarities with all stored innovations and select the most similar innovation based on ranking scores. This approach ensures we capture the most relevant analogous solution at each abstraction level.

This multi-level matching approach offers several key advantages. By considering similarities at different abstraction levels, the system can identify non-obvious connections between problems and potential solutions from disparate domains. The preservation of implementation details alongside abstract principles ensures that identified solutions remain practically applicable rather than purely theoretical. Moreover, the structured format maintains clear traceability between abstract concepts and concrete implementations, facilitating effective knowledge transfer across domains.

Through this comprehensive approach, our system enables the discovery of innovative solutions that might be overlooked by traditional retrieval methods, while ensuring that the identified connections are both meaningful and actionable for practical problem-solving.

\begin{figure}[h]
    \centering
    \includegraphics[width=\textwidth]{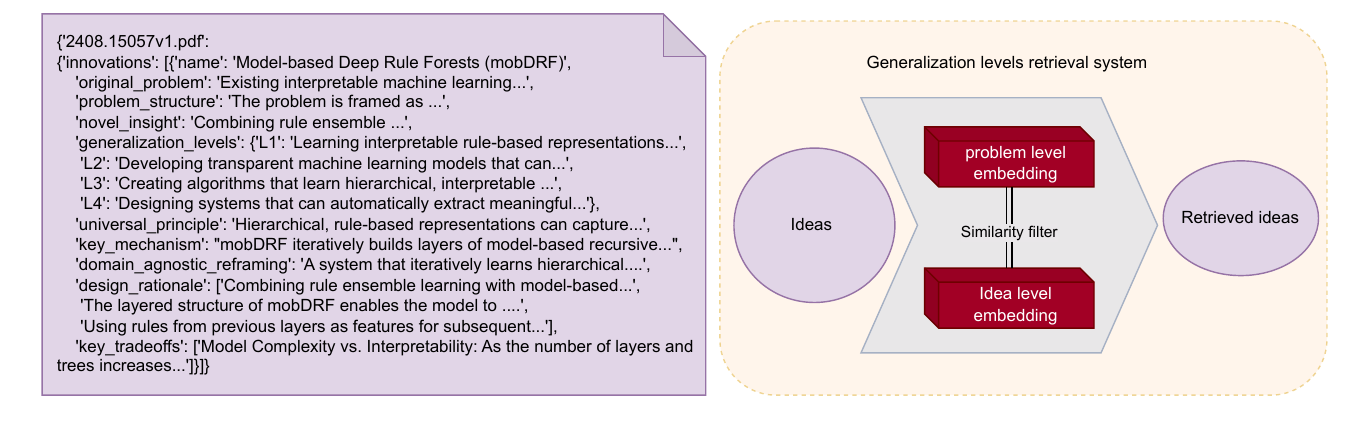} 
    \caption{Semi-strctured idea data format and the level-wise retrieval system}
    \label{fig:example}
\end{figure}
\subsection{Combinatorial Process}
After retrieving relevant innovations through our multi-level matching system, we implement a two-stage combinatorial process that specifically emphasizes combinatorial creativity - the ability to merge existing ideas and concepts in novel ways to generate innovative solutions. This process systematically explores potential combinations of existing innovations while maintaining their practical applicability.

The first stage implements parallel processing across different generalization levels to maximize the potential for novel combinations. For each generalization level, relevant innovations are batched together and analyzed by an AI agent specializing in combinational creativity. The agent examines each innovation through three key perspectives: (1) Component Analysis - breaking down innovations into fundamental mechanisms and principles that could be recombined, (2) Cross-domain Application - identifying how components could be adapted or reinterpreted in new contexts, and (3) Building Block Assessment - evaluating whether components can serve as foundations for new solutions. This structured decomposition is crucial for combinatorial creativity as it creates a rich pool of elements that can be recombined in novel ways.

The second stage integrates insights from all levels to generate cohesive solutions. An integration agent reviews the analyses from the first stage, focusing on feasibility and innovativeness, while considering the relationships between different innovations and their potential contributions to solving the original problem. The agent synthesizes this information to generate solutions characterized by four key aspects: (1) a problem structure that frames how the solution conceptualizes the challenge, (2) a design rationale explaining key implementation decisions, (3) universal principles that capture core ideas applicable across domains, and (4) key mechanisms detailing technical implementation details.

This two-stage process enhances combinatorial creativity in several ways. The parallel level-wise processing in the first stage ensures a diverse pool of components from different abstraction levels, increasing the potential for novel combinations. The structured decomposition helps identify non-obvious connections between components, while the integration stage ensures that these combinations form coherent and practical solutions. By maintaining both breadth in component exploration and depth in integration, the process supports the systematic generation of innovations that bridge multiple domains and approaches.

\section{Experiments}
\subsection{Quantitative Evaluation}
To evaluate our framework's effectiveness in generating innovative ideas through combinatorial creativity, we conducted experiments using the OAG-Bench dataset(\cite{zhang2024oag}). This dataset provides an ideal testing ground as it contains papers along with their important references, allowing us to assess how well our system can generate ideas that align with actual research developments. Since most of these papers were peer-reviewed and published in reputable venues, they inherently satisfy both the novelty and value criteria of creative products - their publication indicates they made original contributions to their fields (novelty) and were deemed scientifically sound and useful by expert reviewers (value). By comparing our generated ideas with these validated research developments, we can evaluate not just similarity, but implicitly assess whether our framework can produce ideas that meet both key criteria for creativity.

\subsubsection{Experimental Setup}
We selected 87 papers from OAG-Bench, each with 3-5 important references identified in their citation network based on citation impact and content relevance. Our evaluation strategy involves using these references as the knowledge base to generate innovative ideas, and then comparing them with the actual innovations presented in the main papers. For each paper, we use our idea extractor agent to extract the core problem statement, which serves as input to our framework. The knowledge base consists of the important references for each paper, structured according to our ideation format. We use Claude-3.5-Sonnet-20241022 as the LLM backend for all idea generation and analysis tasks.

\subsubsection{Baseline and Metrics}
We implemented a baseline system that directly generates ideas given a problem statement, without our multi-level retrieval and combinatorial process. Both our framework and the baseline produce structured outputs in JSON format containing key fields: problem structure, design rationale, universal principle, and key mechanism.

To assess the quality of generated ideas, we compute semantic similarity scores between each field of the generated ideas and the corresponding sections in the target papers. We use allenai-specter model(\cite{specter2020cohan}) to generate embeddings and calculate cosine similarity for each field:
Problem Structure Similarity (PS-Sim),
Design Rationale Similarity (DR-Sim),
Universal Principle Similarity (UP-Sim),
Key Mechanism Similarity (KM-Sim).

\subsubsection{Results}
\begin{figure}[h!]
  \centering
  \includegraphics[ width=\linewidth]{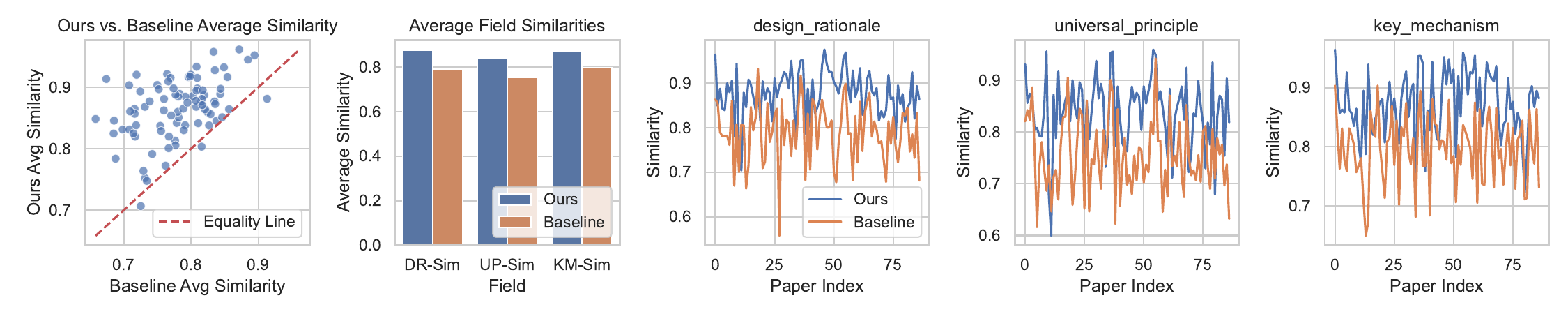}
  \caption{Comparative analysis of similarity scores between our framework and baseline. From left to right: (1) Scatter plot showing overall performance comparison with the equality line, (2) Bar chart comparing average similarities across three key metrics, (3-5) Line plots showing paper-by-paper comparison for design rationale, universal principle, and key mechanism similarities respectively. Our method consistently outperforms the baseline across all metrics, with particularly strong advantages in stability and high-end performance.}
  \label{fig:similarity_comparison}
\end{figure}
Our framework demonstrates consistent superior performance over the baseline across all similarity metrics. As shown in Figure 3, we visualize the comparison results from multiple perspectives.

The scatter plot (leftmost) compares the overall average similarities between our method and the baseline, with most points lying above the equality line, indicating systematic improvement. The bar chart shows the average field similarities, where our method achieves higher scores across all three key metrics: DR-Sim (0.85 vs 0.78), UP-Sim (0.83 vs 0.75), and KM-Sim (0.87 vs 0.77).

The three line plots provide a detailed paper-by-paper comparison for each metric. In design rationale similarity, our method maintains more stable performance with similarity scores consistently above 0.8, while the baseline shows higher variance, occasionally dropping below 0.7. For universal principle similarity, our approach demonstrates particularly strong advantages in maintaining high similarity (mostly above 0.8) compared to the baseline's fluctuating performance. The key mechanism similarity comparison shows the most significant improvement, with our method achieving consistently higher scores around 0.85-0.95 while the baseline frequently drops to 0.7-0.8. These results strongly suggest that Large Language Models are capable of effective combinatorial creativity, successfully combining and adapting existing ideas to generate novel solutions that align well with real research developments.

\subsection{Qualitative Evaluation}
To better understand how our framework generates ideas compared to actual research developments, we conducted a detailed qualitative analysis of the solutions. We present a comparative analysis of three representative cases in Table 1, with the test and target samples provided in Appendix A.

\begin{table*}[h]
\small
\begin{tabular}{p{2.5cm}p{12cm}}
\toprule
\multicolumn{2}{l}{\textbf{Case 1: Adaptive Optimization}} \\
Core Innovation & Both solutions identify the fundamental need to decouple regularization from adaptive updates, demonstrating identical conceptual breakthroughs \\
Technical Design & Near-perfect alignment in proposing a two-step update process, with both separating weight decay from adaptive gradient updates \\
Implementation & Highly similar mathematical formulations, differing only in notation \\
\midrule
\multicolumn{2}{l}{\textbf{Case 2: Text Matching Architecture}} \\
Core Innovation & Remarkable alignment in proposing parallel processing streams for different matching signals, showing identical architectural insights \\
Technical Design & Both solutions independently arrive at a three-stream architecture, demonstrating convergent thinking in component design \\
Implementation & Almost identical component specifications (hierarchical encoding, relevance matching, semantic matching), with minor variations in attention mechanism details \\
\midrule
\multicolumn{2}{l}{\textbf{Case 3: Cold-start Recommendation}} \\
Core Innovation & Strong conceptual alignment in framing the problem as Bayesian active learning with uncertainty modeling \\
Technical Design & Both propose strategic exploration through uncertainty-guided selection, showing identical theoretical foundations \\
Implementation & Highly similar Bayesian update mechanisms, with target solution adding a complementary density weighting factor \\
\bottomrule
\end{tabular}
\caption{Comparison between generated ideas and actual research developments}
\end{table*}

The analysis reveals remarkable alignment between generated ideas and actual research developments. In Case 1, both solutions independently arrive at the crucial insight of separating regularization effects from adaptive updates, with nearly identical mathematical implementations. Case 2 shows even more striking similarity, where both approaches propose exactly the same three-stream architecture with matching component functions. Case 3 demonstrates how our framework can capture sophisticated theoretical frameworks, matching the target solution's Bayesian uncertainty modeling approach.

The observed differences are minimal and primarily in supplementary details rather than core concepts or approaches. For instance, in Case 1, the mathematical notation differs slightly but represents the same underlying mechanism. In Case 2, both solutions propose identical architectural components with minor variations in specific attention mechanisms. Even in Case 3, where the target solution adds density weighting, this represents an optimization of the same fundamental approach rather than a conceptual difference.

This strong alignment across multiple cases suggests that our framework can effectively capture and recombine key insights from existing research to generate solutions that closely parallel actual research developments. The complete samples are provided in Appendix A for detailed reference.

\section{Conclusion}
This paper presents a novel framework for enhancing combinatorial creativity using Large Language Models, specifically designed to facilitate combinatorial creativity for idea generation.The experimental results on the OAG-Bench dataset demonstrate that our framework consistently outperforms baseline approaches in generating ideas that align with real research developments. Particularly strong performance in key mechanism and design rationale similarities suggests that our approach successfully captures both technical depth and logical coherence in the generated ideas. These results indicate that LLMs, when guided by appropriate frameworks, can effectively support combinatorial creativity in research and innovation.

Looking forward, this work opens several promising directions for future research, particularly in deeper integration with computational creativity theory. Following Boden's theoretical framework, our work on combinatorial creativity could be extended to realize exploratory creativity by developing mechanisms for systematically searching conceptual spaces, and transformative creativity by enabling the modification of these spaces' constraints. From the Process perspective, the framework could be enhanced to support different creative reasoning patterns such as analogical thinking and conceptual blending, while incorporating real-time feedback mechanisms to enable more dynamic creative processes.

From the Product perspective, there is a crucial need to develop more comprehensive evaluation frameworks that can assess different types of creativity. While our current evaluation focuses on alignment with existing research developments, future work should develop metrics that can differentiate between P-creativity (ideas novel to the system) and H-creativity (historically novel ideas), as well as assess different levels of creative achievement following Kaufman and Beghetto's "Four C's" model. Additionally, developing benchmarks that can evaluate not just novelty and value, but also influence potential and domain-specific impact measures would significantly advance our understanding of machine creativity. 

Furthermore, the framework could be extended to handle more diverse types of knowledge sources and adapt to specific domain requirements, all while maintaining theoretical grounding in established creativity models. We believe our framework provides a solid foundation for developing more sophisticated tools that bridge computational creativity theory with practical AI-assisted innovation.
\nocite{yang2024qwen2}\nocite{10.1145/3167476}\nocite{achiam2023gpt4}\nocite{Anthropic2024claude3}\nocite{Anthropic2024claude3}\nocite{dubey2024llama3}\nocite{wang2305scimon}\nocite{felin2024theory}
\bibliographystyle{unsrtnat}
\bibliography{references}  






\appendix
\section{Appendix}
\label{appendix:samples}

This appendix presents the complete comparison samples between generated solutions (Test) and actual research outcomes (Target) for all 87 papers in our evaluation dataset.

\begin{longtable}{p{0.48\textwidth}p{0.48\textwidth}}
\caption{Complete comparison samples for all test cases} \\
\toprule
\textbf{Generated Solution (Test)} & \textbf{Actual Research (Target)} \\
\midrule
\endfirsthead

\multicolumn{2}{c}{Table continues from previous page} \\
\toprule
\textbf{Generated Solution (Test)} & \textbf{Actual Research (Target)} \\
\midrule
\endhead

\midrule
\multicolumn{2}{r}{Continues on next page} \\
\endfoot

\bottomrule
\endlastfoot

\multicolumn{2}{l}{\textbf{Case 1}} \\
\textit{Problem Structure:} & \textit{Problem Structure:} \\
The problem is structured as a separation of concerns between adaptive gradient updates and regularization effects. By treating these as independent processes, the solution ensures that regularization remains consistent across all parameters regardless of their gradient histories. & The problem is framed as a need to separate weight regularization from the gradient-based update step in adaptive optimization methods. The structure recognizes that the interaction between adaptive gradients and regularization requires different treatment than in standard SGD. \\
\textit{Design Rationale:} & \textit{Design Rationale:} \\
Separating the weight decay step from adaptive updates prevents the regularization from being scaled by adaptive learning rates, ensuring uniform regularization pressure Using a fixed learning rate for regularization provides direct control over regularization strength independent of gradient-based optimization The two-step update process maintains the benefits of adaptive methods while fixing their regularization issues & Separating weight decay from gradient updates prevents the adaptive scaling of gradients from affecting the regularization strength, ensuring consistent regularization across all parameters regardless of their gradient history This separation makes the weight decay effect more predictable and controllable, as it's no longer dependent on the complex dynamics of adaptive gradient scaling The decoupled approach allows for more direct control over the trade-off between optimization and regularization, as their effects are now independent \\
\textit{Universal Principle:} & \textit{Universal Principle:} \\
Complex optimization processes can be improved by decoupling interdependent components that serve different purposes, allowing each component to operate independently and more effectively. & When multiple adaptive mechanisms interact through shared parameters, explicitly separating their effects can provide better control and improved outcomes than combining them into a single mechanism. \\
\textit{Key Mechanism:} & \textit{Key Mechanism:} \\
The update process occurs in two distinct steps: 1) Apply adaptive gradient update: $\theta$t' = $\theta$t-1 - $\alpha·mt/(\sqrt{v}t + \epsilon)$, where mt and vt are the Adam moment estimates. 2) Apply fixed weight decay: $\theta$t = $\theta$t' - $\lambda$·$\theta$t', where $\lambda$ is a fixed regularization rate. This ensures consistent regularization across all parameters while preserving adaptive gradient benefits. & Weight decay is applied separately from the adaptive gradient update: $\theta t+1 = (1-\lambda)\theta t - \alpha Mt \delta ft(\theta t)$, where $\lambda$ is the weight decay rate, $\alpha$ is the learning rate, Mt is the adaptive learning rate matrix, and $\delta ft(\theta t)$ is the gradient. This contrasts with L2 regularization which adds $\lambda'||\theta||^2$ to the loss function. \\

\midrule
\multicolumn{2}{l}{\textbf{Case 2}} \\
\textit{Problem Structure:} & \textit{Problem Structure:} \\
The problem is structured as a parallel processing challenge where different types of matching signals (relevance and semantic) need to be captured and integrated at multiple granularities. This approach recognizes that text similarity is multifaceted and requires simultaneous processing of different matching paradigms. & The problem is structured as a need to bridge two complementary approaches to text matching - relevance matching that focuses on term/keyword matching with importance weighting, and semantic matching that captures deeper contextual meaning. This requires a unified architecture that can effectively combine both types of signals while preserving their distinct strengths. \\
\textit{Design Rationale:} & \textit{Design Rationale:} \\
The hierarchical processing at multiple granularities (character, word, phrase, sentence) ensures comprehensive capture of similarity signals at all levels of text structure Parallel processing streams for relevance and semantic matching allow each component to optimize for its specific objective while sharing underlying representations The weighted integration mechanism enables dynamic balancing of different matching signals based on input characteristics and task requirements & Using parallel processing streams allows the model to simultaneously capture both lexical/surface matches and semantic relationships without one interfering with the other The hybrid encoder approach provides flexibility in representation learning, allowing the model to capture patterns at multiple granularities Integration of signals at the final classification stage allows the model to learn optimal combinations for different types of tasks \\
\textit{Universal Principle:} & \textit{Universal Principle:} \\
Complex comparison tasks can be effectively solved by decomposing them into parallel specialized components that operate at multiple scales, with dynamic integration of their outputs. & Complex information processing tasks often require both surface-level pattern matching and deeper semantic understanding, which can be effectively combined through parallel processing streams that maintain their distinct characteristics. \\
\textit{Key Mechanism:} & \textit{Key Mechanism:} \\
The solution employs three parallel streams: 1) A hierarchical encoder using character-CNN, word-LSTM, and sentence-Transformer layers, 2) A relevance matching component using histogram-based representations with cascade k-max pooling, and 3) A semantic matching component using bi-directional attention flow. These streams are integrated through a learned weighting mechanism that combines their signals based on input characteristics. & HCAN comprises three main components working in parallel: 1) A hybrid encoder module using ConvNet and LSTM-based encoders to obtain contextual representations at multiple granularities, 2) A relevance matching module that measures soft term matches with importance weighting, 3) A semantic matching module with co-attention mechanisms for context-aware semantic understanding. The outputs from all components are integrated through a final classification layer. \\
\midrule
\multicolumn{2}{l}{\textbf{Case 3}} \\
\textit{Problem Structure:} & \textit{Problem Structure:} \\
The solution frames the cold-start problem as a Bayesian active learning challenge where uncertainty in user preferences drives the exploration strategy. It combines uncertainty modeling through dropout-based Bayesian approximation with strategic article selection to efficiently build user preference profiles. & The problem is structured as an active learning challenge where the system must efficiently learn user preferences through minimal strategic interactions, while maintaining personalization capabilities through Bayesian uncertainty modeling and one-shot learning adaptation. \\
\textit{Design Rationale:} & \textit{Design Rationale:} \\
Using dropout-based Bayesian approximation provides a computationally efficient way to model uncertainty in user preferences without requiring complex probabilistic models The two-stage selection process (topic-level followed by article-level) enables efficient exploration of the preference space while maintaining reasonable computational complexity Incorporating semantic embeddings allows for transfer learning and better generalization from limited user feedback & Bayesian modeling captures uncertainty in user preferences, which is essential for making informed decisions about which articles to query Density weighting helps avoid querying outlier examples that may be informative but not representative of the typical content space Combining one-shot learning with active querying enables rapid adaptation to individual users while minimizing the amount of feedback required \\
\textit{Universal Principle:} & \textit{Universal Principle:} \\
Uncertainty-driven exploration combined with hierarchical decomposition can efficiently solve cold-start problems in high-dimensional spaces with limited feedback & Efficient learning of complex preferences can be achieved by strategically selecting a small number of maximally informative examples through uncertainty-guided active learning. \\
\textit{Key Mechanism:} & \textit{Key Mechanism:} \\
Maintains a user preference vector with dropout-based uncertainty estimation. Uses Monte Carlo sampling with dropout to generate multiple preference estimates. Selects articles through a two-stage process: first choosing topics with highest uncertainty, then selecting specific articles within those topics using expected information gain. Updates preference vector using Bayesian updates based on user feedback. & POLAR++ combines three key components: 1) A Bayesian neural network to model uncertainty in user preferences and enable active learning, 2) An attention-based CNN for measuring article similarity, and 3) A density-weighted Expected Loss Optimization strategy for selecting articles to query. The active learning component selects articles that maximize expected loss while considering density to avoid outliers: $arg max_di (EL(di|dq,S) * 1/|U| \sum c(di,dj))$, where EL is expected loss and c is similarity. \\
\midrule
\multicolumn{2}{l}{\textbf{Case 4}} \\
\textit{Problem Structure:} & \textit{Problem Structure:} \\
The problem is framed as replacing sequential RNN processing with a parallel architecture that can effectively capture both local and global text relationships while maintaining or improving comprehension accuracy. The solution uses a hierarchical approach combining multiple levels of parallel processing. & The problem is framed as designing a fully feed-forward architecture that can capture both local and global text interactions through parallel computation, while maintaining the ability to model complex relationships needed for reading comprehension. \\
\textit{Design Rationale:} & \textit{Design Rationale:} \\
The use of self-attention mechanisms enables fully parallel processing of input text, dramatically improving computational efficiency compared to RNNs A multi-level hierarchical structure allows the model to capture relationships at different granularities simultaneously, maintaining comprehensive text understanding Integration of positional encodings and residual connections preserves sequential information while enabling parallel processing & The combination of convolutions and self-attention provides complementary modeling capabilities - convolutions capture local patterns efficiently while self-attention handles long-range dependencies Removing recurrent connections allows the model to process all positions in parallel, dramatically improving computational efficiency Using residual connections and layer normalization helps train deep stacks of these blocks effectively by providing direct gradient paths and normalizing activations \\
\textit{Universal Principle:} & \textit{Universal Principle:} \\
Complex sequential processes can be parallelized by decomposing them into multiple levels of simultaneous operations while maintaining information flow between levels. & Complex relationships in sequential data can be effectively modeled by decomposing them into local and global components that can be processed in parallel. \\
\textit{Key Mechanism:} & \textit{Key Mechanism:} \\
The architecture consists of multiple parallel processing streams: character-level CNNs process subword information, while transformer layers with self-attention handle word and sentence-level relationships. Positional encodings maintain sequence information, and multi-head attention allows different heads to focus on different relationship types. The model processes all positions simultaneously through multiple layers of self-attention and feed-forward networks. & The model uses a stack of encoder blocks, where each block consists of: 1) Positional encoding, 2) Depth-wise separable convolutions to capture local structure, 3) Multi-head self-attention to model global interactions, 4) Feed-forward layers. Each component is wrapped in residual connections and layer normalization. The architecture eliminates recurrent connections entirely in favor of purely feed-forward operations. \\
\midrule
\multicolumn{2}{l}{\textbf{Case 5}} \\
\textit{Problem Structure:} & \textit{Problem Structure:} \\
The problem is structured as a hierarchical natural language inference task where both semantic meaning and structural relationships need to be captured effectively. The solution addresses this by implementing a tree-based architecture that processes information at multiple levels while maintaining computational efficiency. & The problem is framed as a three-stage process: input encoding using bidirectional LSTM, local inference modeling through soft attention alignment, and inference composition that combines local inference information to make the final prediction. \\
\textit{Design Rationale:} & \textit{Design Rationale:} \\
Tree-LSTM structure allows for better capture of hierarchical relationships in language compared to sequential models, while maintaining the ability to handle variable-length inputs Structured attention mechanisms enable more nuanced comparison between premise and hypothesis by considering both local and global relationships Gated information flow controls help maintain model interpretability while allowing selective information propagation through the hierarchy & Using BiLSTM for input encoding allows capturing contextual information effectively while maintaining computational efficiency compared to more complex architectures Local inference enhancement through difference and element-wise product helps capture contradiction and similarity relationships more explicitly Combining average and max pooling in inference composition helps capture both overall patterns and the most significant local inference information \\
\textit{Universal Principle:} & \textit{Universal Principle:} \\
Complex hierarchical relationships can be effectively modeled through tree-structured architectures that combine local and global information processing with selective information flow. & Complex relational understanding can be achieved through a structured process of alignment, local comparison, and global composition of information. \\
\textit{Key Mechanism:} & \textit{Key Mechanism:} \\
The solution uses a Tree-LSTM encoder to process parsed sentence structures, implements structured attention for comparing premise and hypothesis at multiple levels, and employs gated mechanisms for selective information flow. The process follows: Input → Parse Tree → Tree-LSTM Encoding → Structured Attention → Gated Composition → Classification & The model consists of three main components: 1) BiLSTM encoding of input sequences, 2) soft alignment between sequences using attention mechanisms with element-wise comparison operations (difference and product), and 3) another BiLSTM layer to compose the local inference information with average/max pooling for final classification. \\
\midrule
\multicolumn{2}{l}{\textbf{Case 6}} \\
\textit{Problem Structure:} & \textit{Problem Structure:} \\
The problem is framed as a trade-off between memory efficiency and model expressiveness, where the goal is to maximize feature representation capability while maintaining minimal memory footprint. The solution uses a dynamic structure that adapts its computational resources based on input complexity. & The problem is framed as finding an architecture that maximizes information flow through narrow bottleneck layers while maintaining computational efficiency. This requires rethinking how residual connections are used with respect to expanded and compressed representations. \\
\textit{Design Rationale:} & \textit{Design Rationale:} \\
The inverted residual structure with bottleneck layers minimizes memory usage during most of the computation while allowing for expressive feature extraction in the expansion phase Dynamic width adjustment enables efficient resource utilization by adapting the network's capacity based on input complexity Integration of channel shuffle operations enhances information flow across groups without significant memory overhead & By connecting bottleneck layers directly, the network maintains a clear separation between capacity (bottleneck width) and expressiveness (expansion ratio), allowing independent optimization of these aspects. The inverted design is more memory efficient since activations of bottleneck layers that need to be stored for residual connections are smaller. This structure allows the network to maintain essential information in the narrow layers while using expansion layers purely for non-linear transformation capability. \\
\textit{Universal Principle:} & \textit{Universal Principle:} \\
Resource efficiency can be achieved through dynamic allocation and transformation of computational capacity, where system expressiveness is maintained through temporary expansion while keeping persistent memory usage low & When a system alternates between compressed and expanded states, connecting the compressed states directly can preserve essential information more efficiently than connecting expanded states. \\
\textit{Key Mechanism:} & \textit{Key Mechanism:} \\
The block uses a thin bottleneck layer (1/4 to 1/6 of expansion size) for input, dynamically expands channels using 1x1 convolution, applies depthwise separable convolution with channel shuffle for spatial filtering, then projects back to a thin representation. The expansion ratio is determined adaptively based on input complexity and available resources. & The inverted residual block first expands the input channels by a factor t (typically 6), applies a depthwise convolution with ReLU6 for non-linear transformation, then projects back to a low-dimensional representation. The skip connection connects these low-dimensional (bottleneck) representations directly. The block can be expressed as F(x) = ANB(x), where A and B are linear transformations and N is a non-linear transformation. \\
\midrule

\multicolumn{2}{l}{\textbf{Case 8}} \\
\textit{Problem Structure:} & \textit{Problem Structure:} \\
The problem is structured as a two-phase learning approach that first separates universal patterns from source-specific biases through parameter transformation, then dynamically combines knowledge from different sources using context-aware attention mechanisms. & The problem is structured as a two-phase approach: (1) learning source-specific representations that capture unique annotation patterns and biases of each source while sharing common knowledge, and (2) dynamically weighting and combining these source representations based on input context to reach an optimal consensus for prediction. \\
\textit{Design Rationale:} & \textit{Design Rationale:} \\
Using parameter transformation in multi-task learning allows the model to effectively separate shared knowledge from source-specific patterns while maintaining computational efficiency The context-aware attention mechanism enables flexible aggregation of source knowledge based on input characteristics, improving model adaptability Integration of source reliability estimation helps handle noisy and inconsistent annotations while maintaining robustness & By decoupling source-specific biases as transformations on a shared base model, the framework can effectively capture both common knowledge and unique patterns of each source Using context-aware attention for dynamic source selection allows the model to adaptively leverage the most relevant sources for each input, rather than relying on static source weights The two-phase structure enables efficient training - first learning robust source representations, then learning how to optimally combine them \\
\textit{Universal Principle:} & \textit{Universal Principle:} \\
Complex systems with multiple information sources can be effectively managed by separating universal patterns from source-specific variations, then dynamically combining them based on context-specific relevance. & When combining multiple information sources with varying reliability/expertise, the relative importance of each source should be determined dynamically based on the specific context rather than using static weights. \\
\textit{Key Mechanism:} & \textit{Key Mechanism:} \\
The solution employs a shared LSTM encoder with source-specific parameter transformations, followed by an attention mechanism that weights source contributions based on input context. The model is trained using a multi-task objective with a noise-aware loss function that considers source reliability estimates. & Uses a two-phase approach: (1) Decoupling Phase - learns source-specific transformation matrices A(k) that capture annotation biases while sharing base model parameters. Each source's unique patterns are modeled as transformations on emission and transition scores. (2) Aggregation Phase - learns an attention module that takes sentence embeddings as input and outputs normalized weights for each source, dynamically combining source-specific matrices into a consensus representation A* for prediction. \\
\midrule


\multicolumn{2}{l}{\textbf{Case 11}} \\
\textit{Problem Structure:} & \textit{Problem Structure:} \\
The problem is structured as an iterative refinement process where Probabilistic Soft Logic (PSL) and complex number embeddings work together to simultaneously handle noise removal and fact inference. PSL provides the logical consistency framework while complex embeddings capture rich relationship patterns. & The problem is structured as an iterative co-training framework where two complementary approaches - rule-based (PSL-KGI) and embedding-based methods - augment each other's strengths. PSL-KGI provides high-quality type predictions while embeddings capture implicit relationships, with feedback loops between them to progressively improve both. \\
\textit{Design Rationale:} & \textit{Design Rationale:} \\
Complex number embeddings naturally handle both symmetric and antisymmetric relations, providing a more expressive representation for different relationship types PSL's continuous [0,1] truth values allow for nuanced uncertainty handling and smooth integration with embedding-based confidence scores The iterative refinement process allows each component to compensate for the other's weaknesses while building on their respective strengths & The iterative feedback loop allows each method to benefit from the other's strengths while mitigating their individual weaknesses Using thresholds to control feedback volume prevents error propagation while maintaining computational feasibility Explicit type supervision from PSL-KGI helps embeddings maintain type compatibility, leading to higher quality predictions \\
\textit{Universal Principle:} & \textit{Universal Principle:} \\
Complex systems can be effectively refined through the iterative application of complementary approaches, where each approach's strengths compensate for the other's weaknesses. & Complementary methods can be combined through iterative feedback loops where each method's outputs become inputs to improve the other method's performance. \\
\textit{Key Mechanism:} & \textit{Key Mechanism:} \\
The framework operates in alternating phases: 1) Complex embeddings learn representations using Hermitian dot product scoring $$\phi(r,s,o) = Re(\langle w_r,e_s,\bar{e}_o\rangle)$$. 2) PSL applies weighted logical rules as constraints, converting them into a convex optimization problem. 3) Results from each phase inform the other through confidence scores and fact filtering. The process repeats until convergence. & 1. PSL-KGI uses ontological rules to predict entity types and relations 2. These predictions train type-aware embeddings (TypeE-X) 3. High-confidence embedding predictions feed back to PSL-KGI as additional evidence 4. Process repeats iteratively with thresholds t2 and t3 controlling feedback volume \\
\midrule
\multicolumn{2}{l}{\textbf{Case 12}} \\
\textit{Problem Structure:} & \textit{Problem Structure:} \\
The problem is structured as a multi-level information extraction task that requires detecting algorithm mentions, understanding their relationships, and constructing a temporal evolution graph. The solution uses a hierarchical approach that processes papers at different granularities to capture both local and global context. & The problem is structured as a multi-level relation extraction task that needs to: (1) identify relevant algorithm mentions across sentence boundaries, (2) model interactions between all abbreviations in the context, (3) jointly predict both the relation between algorithm pairs and their types, while leveraging weak supervision signals. \\
\textit{Design Rationale:} & \textit{Design Rationale:} \\
Combining transformer architecture with PCNNs allows the system to capture both long-range dependencies and local features around algorithm mentions, making it effective at understanding complex relationships. Using interpretable feature representations ensures that the resulting algorithm graph is not just accurate but also meaningful and useful to researchers. Incorporating temporal information through both publication dates and in-text cues helps establish reliable evolutionary paths between algorithms. & The combination of local and global context allows capturing both explicit comparisons within sentences and implicit comparisons across sentences The abbreviation-attention mechanism helps disambiguate entity types and capture implicit relationships by modeling interactions between all similar entities Joint prediction of relations and types provides mutual reinforcement and helps filter out irrelevant entities \\
\textit{Universal Principle:} & \textit{Universal Principle:} \\
Complex evolutionary relationships in any domain can be understood by combining local feature analysis with global context while maintaining interpretability through human-readable representations. & Complex relationships between entities can be better understood by considering both their immediate context and broader interactions with similar entities in the surrounding discourse. \\
\textit{Key Mechanism:} & \textit{Key Mechanism:} \\
The system uses a transformer backbone to process entire papers, with PCNNs handling local context around algorithm mentions. Bi-affine attention scores relationships between algorithms, while temporal embeddings ensure chronological consistency. The final graph is constructed using weighted edges based on relationship scores and temporal ordering. & CANTOR combines: (1) A single-sentence module using PCNN for local context, (2) A cross-sentence module with self-attention for paragraph-level context and abbreviation-attention for modeling interactions between all abbreviations, (3) Joint prediction of relations and entity types, (4) Interpolation between single and cross-sentence predictions with learned weights $\lambda$1 and $\lambda$2. \\
\midrule


\multicolumn{2}{l}{\textbf{Case 15}} \\
\textit{Problem Structure:} & \textit{Problem Structure:} \\
The problem is structured as a multi-level feature processing challenge where different types of information (structural, temporal, content) need to be processed at different time scales while maintaining their interactions. The solution uses a hierarchical architecture that processes features at different levels while maintaining global context. & The problem is structured as a need to create a model that can process three types of information: (1) node-level content and structural features through graph convolution, (2) short-term temporal patterns through attention over recent states, and (3) long-term dependencies through recurrent processing - all while maintaining the ability to detect anomalies in an online fashion. \\
\textit{Design Rationale:} & \textit{Design Rationale:} \\
The hierarchical attention mechanism allows the model to capture both short-term behaviors and long-term patterns while maintaining computational efficiency The global context memory provides a way to maintain and utilize long-term patterns while processing current information The GRU-based processing enables selective updating of information, allowing the model to maintain relevant historical information while incorporating new patterns & The combination of GCN, attention and GRU allows capturing complementary aspects of the data - GCN for structural patterns, attention for recent behavioral changes, and GRU for long-term dependencies Using attention over a time window provides flexibility in capturing short-term patterns while avoiding the limitations of fixed-size contexts The end-to-end architecture allows the different components to adapt to each other during training, leading to more effective representations for anomaly detection \\
\textit{Universal Principle:} & \textit{Universal Principle:} \\
Complex temporal-structural patterns can be effectively captured through a hierarchical processing system that maintains global context while processing local features at multiple scales & Complex temporal patterns in dynamic systems can be better understood by explicitly modeling and combining information at different time scales using specialized architectural components. \\
\textit{Key Mechanism:} & \textit{Key Mechanism:} \\
The solution implements a two-level hierarchical attention mechanism with a global context memory. The lower level processes local structural and temporal features using graph convolutions and attention mechanisms, while the higher level captures long-term patterns and global structure. A GRU-based component selectively updates node representations based on new information and global context. Anomalies are detected by comparing current states with expected patterns derived from the global context. & The framework processes each timestep through three stages: (1) GCN layer processes structural and content features of current snapshot, (2) Attention mechanism aggregates states from recent time window to capture short-term patterns, (3) GRU combines current state, windowed context, and previous hidden state to maintain long-term memory. The outputs are node embeddings that can be used to score potential anomalies. \\
\midrule




\multicolumn{2}{l}{\textbf{Case 20}} \\
\textit{Problem Structure:} & \textit{Problem Structure:} \\
The problem is structured as a multi-stream feature extraction and fusion challenge, where each modality (visual, audio, text) requires specialized processing before integration. The focus is on extracting stable, meaningful features from each channel while accounting for noise and irrelevant variations. & The problem is framed as a multi-level feature extraction and fusion challenge, where different modalities (visual, audio, text) capture complementary aspects of deceptive behavior. The approach uses both low-level features directly from the data and high-level semantic features derived from micro-expression detection. \\
\textit{Design Rationale:} & \textit{Design Rationale:} \\
Camera-stabilized dense trajectory extraction provides robust visual features by separating genuine behavioral cues from camera motion artifacts Fisher Vector encoding creates fixed-length representations that capture rich statistical information about feature distributions across all modalities Late fusion with attention mechanisms allows the system to dynamically weight different modalities based on their reliability and relevance & Multiple modalities provide complementary information that helps capture different aspects of deceptive behavior, making the system more robust than single-modality approaches Using both low-level and high-level features allows the system to capture both subtle unconscious patterns and semantic behavioral cues, providing a more complete picture of the phenomenon Late fusion of different classifiers allows each modality to be processed optimally while still benefiting from their complementary strengths \\
\textit{Universal Principle:} & \textit{Universal Principle:} \\
Complex behavioral patterns can be accurately detected by combining multiple streams of complementary information through hierarchical feature extraction and adaptive fusion mechanisms. & Complex behavioral phenomena can be better understood by analyzing multiple information streams at different levels of abstraction and combining them systematically. \\
\textit{Key Mechanism:} & \textit{Key Mechanism:} \\
The system first extracts dense trajectories from video using SURF features and optical flow, stabilized through RANSAC-based camera motion estimation. Audio and text features are processed through specialized extractors. All features are encoded into fixed-length representations using Fisher Vectors. Finally, an attention-based late fusion mechanism combines evidence across modalities for final classification. & The system extracts features from three modalities (video, audio, text) using IDT for motion, MFCC for audio, and GloVe for text. It then builds two levels of features - low-level features encoded using Fisher Vectors and high-level semantic features from micro-expression detection. These features are processed by separate classifiers and combined through late fusion with learned weights. \\
\midrule
\multicolumn{2}{l}{\textbf{Case 21}} \\
\textit{Problem Structure:} & \textit{Problem Structure:} \\
The problem is approached as a multi-layered challenge requiring the integration of linguistic, visual, and contextual analysis of court trial data. The framework addresses both the data collection and validation aspects while maintaining ethical considerations and scientific rigor. & The problem is framed as a multimodal classification task that combines verbal and non-verbal behavioral cues extracted from authentic trial videos to distinguish between truthful and deceptive testimony. This structure assumes that deceptive behavior manifests through multiple channels and that real high-stakes scenarios provide more reliable deception indicators compared to simulated settings. \\
\textit{Design Rationale:} & \textit{Design Rationale:} \\
Combining multiple modalities (linguistic, visual, physiological) provides redundant and complementary signals that make deception detection more robust in high-stakes scenarios Using a two-stage verification process (automated analysis followed by expert review) balances efficiency with accuracy Implementing standardized annotation procedures ensures consistency and reproducibility of findings & Real trial data provides more reliable deception indicators compared to artificial settings since the stakes and motivations are authentic, leading to more genuine behavioral manifestations Combining multiple modalities (verbal + non-verbal) allows capturing complementary aspects of deceptive behavior that may not be evident from a single channel Using standardized coding schemes for non-verbal behavior annotation enables systematic analysis of behavioral patterns \\
\textit{Universal Principle:} & \textit{Universal Principle:} \\
Complex human behaviors can be more accurately assessed through the systematic integration of multiple information streams combined with expert validation & The authenticity of human behavior can be more accurately assessed by jointly analyzing multiple behavioral channels in real high-stakes contexts rather than artificial settings. \\
\textit{Key Mechanism:} & \textit{Key Mechanism:} \\
The system first extracts multimodal features (stylometric, visual, audio) from court trial videos, applies machine learning classifiers to identify potential deception, and validates results through expert review. The process uses adapted MUMIN annotation schemes for systematic documentation and includes privacy-preserving protocols. & The system extracts and combines verbal features (unigrams, bigrams from transcripts) and non-verbal features (facial expressions, hand gestures manually annotated using MUMIN coding scheme) from trial videos. These features are used to train machine learning classifiers (Decision Trees, Random Forest) to discriminate between truthful and deceptive testimony. \\
\midrule
\multicolumn{2}{l}{\textbf{Case 22}} \\
\textit{Problem Structure:} & \textit{Problem Structure:} \\
The problem is framed as a multi-modal classification challenge that requires simultaneous analysis of verbal, non-verbal, and physiological signals from genuine deception instances. The solution recognizes that reliable deception detection must adapt to individual baselines and contextual factors while integrating multiple channels of human communication. & The problem is structured as a need to create a comprehensive dataset that captures authentic deceptive behavior across multiple modalities (verbal and non-verbal) from real-life high-stakes scenarios where ground truth about deception can be verified. \\
\textit{Design Rationale:} & \textit{Design Rationale:} \\
Integration of multiple modalities (thermal, linguistic, physiological) provides redundancy and cross-validation, making the system more robust against individual attempts to mask deceptive behaviors Adaptive baseline normalization accounts for individual differences and contextual factors, improving accuracy across different scenarios and populations Use of non-invasive sensors and privacy-preserving techniques makes the system practical for real-world deployment & Real high-stakes scenarios provide authentic emotional arousal and motivation that is difficult to replicate in lab settings, leading to more genuine behavioral patterns Using multiple sources (trials and interviews) helps capture varied contexts while maintaining ability to verify ground truth through outcomes Detailed manual annotation of multiple modalities enables comprehensive analysis of how deception manifests across verbal and non-verbal channels \\
\textit{Universal Principle:} & \textit{Universal Principle:} \\
Complex human behaviors can be more accurately detected and classified through the simultaneous analysis of multiple communication channels, normalized against individual baselines and contextual factors. & High-stakes real-world scenarios where outcomes can be verified provide more reliable data for studying authentic human behavior compared to artificial experimental settings. \\
\textit{Key Mechanism:} & \textit{Key Mechanism:} \\
The system combines thermal imaging, linguistic analysis, and physiological measurements through a three-stage process: 1) Multi-modal feature extraction using non-invasive sensors, 2) Individual baseline normalization and contextual weighting of features, 3) Machine learning classification using concatenated feature vectors. Features are processed through parallel pipelines before being integrated for final classification. & Collection of video clips from public trials and interviews where deceptive/truthful nature can be verified through verdicts, evidence or confession. Videos are manually transcribed and annotated for verbal and non-verbal behaviors using established coding schemes. Multiple quality controls ensure clear visibility of subjects and high audio quality. \\
\midrule
\multicolumn{2}{l}{\textbf{Case 23}} \\
\textit{Problem Structure:} & \textit{Problem Structure:} \\
The problem is reframed as a reading comprehension task where entity extraction becomes a process of answering specific queries about entity types. This unified approach eliminates the traditional distinction between flat and nested entity recognition by treating all entity extraction as query-response pairs. & The problem is reframed as extracting answer spans from text in response to natural language queries about specific entity types, rather than as sequence labeling. Each entity type becomes a question about what entities of that type are mentioned in the text. \\
\textit{Design Rationale:} & \textit{Design Rationale:} \\
Using natural language queries provides flexibility to handle any entity type without architectural changes, making the system extensible and maintainable The multi-turn approach allows for explicit modeling of hierarchical relationships between entities, enabling natural handling of nested structures Query-based BMEO tagging enables precise boundary detection while maintaining the ability to identify multiple overlapping entities & Converting entity types to questions allows incorporation of rich semantic knowledge about what to extract, going beyond simple label indices Having separate queries for each entity type naturally handles overlapping entities since each extraction is independent Using binary classifiers for start/end positions rather than n-way classification allows extracting multiple spans per query \\
\textit{Universal Principle:} & \textit{Universal Principle:} \\
Complex hierarchical classification tasks can be transformed into a series of simpler query-response interactions, allowing unified handling of different complexity levels within the same framework & Complex information extraction tasks can be decomposed into simpler question-answering sub-tasks that naturally handle overlapping elements and incorporate domain knowledge. \\
\textit{Key Mechanism:} & \textit{Key Mechanism:} \\
The system processes text through a BERT-based reading comprehension model that performs query-based BMEO tagging. For each entity type query, it predicts B(egin), M(iddle), E(nd), or O(utside) tags for each token. Nested entities are handled through multi-turn processing where initial entity extractions inform subsequent queries. & Each entity type is associated with a natural language query describing what to extract. A BERT-based MRC model processes both query and text to identify relevant spans. Multiple entity types are handled through separate queries, naturally allowing for overlapping entities. The model uses binary classifiers to identify start/end positions of spans that answer each query. \\
\midrule
\multicolumn{2}{l}{\textbf{Case 24}} \\
\textit{Problem Structure:} & \textit{Problem Structure:} \\
The problem is framed as a need to simultaneously capture fine-grained spatial details and global contextual information in aerial imagery segmentation. The solution addresses this through a multi-scale architecture that dynamically weights feature importance at different scales while maintaining spatial precision. & The problem is structured as a need to simultaneously capture both global context and local details while maintaining efficient information flow across different scales. This requires balancing the preservation of fine spatial details with the extraction of broader contextual features. \\
\textit{Design Rationale:} & \textit{Design Rationale:} \\
Integration of attention mechanisms allows the network to dynamically focus on relevant features at each scale, improving the model's ability to handle complex urban structures Multi-scale feature processing with parallel atrous convolutions enables capture of context at various receptive fields without losing resolution Enhanced skip connections with feature refinement modules ensure effective information flow between encoder and decoder paths & The PPL module with eight different pooling sizes enables extraction of features at multiple scales, allowing the model to capture both local patterns and global context simultaneously Skip connections between corresponding encoder and decoder layers preserve fine spatial details that would otherwise be lost during downsampling, crucial for accurate boundary detection The combination of U-Net architecture with PPL creates multiple paths for information flow, enabling the model to better handle complex spatial relationships and overlapping features \\
\textit{Universal Principle:} & \textit{Universal Principle:} \\
Complex hierarchical patterns can be effectively captured by processing information at multiple scales simultaneously while using attention mechanisms to dynamically weight the importance of different scales based on context. & Parallel processing of information at multiple scales with bidirectional information flow enables systems to simultaneously capture both broad patterns and fine details. \\
\textit{Key Mechanism:} & \textit{Key Mechanism:} \\
The architecture employs parallel atrous convolutions with different dilation rates at each encoder level, combined with scale-specific attention modules. Features from different scales are fused using learnable weights, and skip connections are enhanced with 1x1 convolutions for channel reduction and 3x3 convolutions for feature refinement. The decoder includes progressive upsampling with multi-scale feature aggregation. & The architecture combines U-Net's encoder-decoder structure with an eight-level pyramid pooling module that: 1. Uses different pooling sizes (1,3,8,9,11,12,13,14) to extract features at multiple scales 2. Concatenates features from different pyramid levels and upsamples them  3. Integrates skip connections to preserve fine spatial details 4. Uses batch normalization to prevent overfitting \\
\midrule
\multicolumn{2}{l}{\textbf{Case 25}} \\
\textit{Problem Structure:} & \textit{Problem Structure:} \\
The problem is framed as creating an efficient knowledge transfer mechanism that preserves essential ranking relationships while allowing architectural flexibility. It focuses on distilling ranking knowledge through a shared metric space that captures relative ordering rather than absolute scores. & The problem is structured as knowledge transfer from a large teacher model to a smaller student model in the context of ranking, where the focus is on learning relative ordering of items rather than class labels. The student model learns from both labeled training data and top-K ranked unlabeled items from the teacher model. \\
\textit{Design Rationale:} & \textit{Design Rationale:} \\
Using a shared metric space allows for model-agnostic knowledge transfer while preserving the geometric relationships that define rankings Implementing rank-invariant factor decomposition enables efficient compression of ranking knowledge while maintaining critical ordering information Adaptive sampling ensures focus on the most informative ranking examples, optimizing the knowledge transfer process & Using only top-K ranked items from teacher focuses the knowledge transfer on the most relevant relationships while avoiding noise from lower-ranked items Dynamic weighting based on position and ranking discrepancy allows adaptive emphasis on items where the student needs most improvement Combining supervision from both ground truth data and teacher predictions provides complementary learning signals \\
\textit{Universal Principle:} & \textit{Universal Principle:} \\
Complex hierarchical relationships can be preserved through geometric representations in a shared space, allowing efficient transfer of structural knowledge between systems of different capacities. & Complex relational knowledge can be effectively transferred between systems of different capacities by focusing on the most important/relevant relationships rather than trying to transfer all relationships. \\
\textit{Key Mechanism:} & \textit{Key Mechanism:} \\
Creates a shared low-dimensional embedding space where both teacher and student models project their ranking outputs. Uses rank-invariant factor decomposition to compress ranking knowledge, followed by adaptive sampling to focus on crucial ranking relationships. The student model is trained using a combination of distillation loss (measuring embedding similarity) and ranking loss (preserving relative ordering). & A student model is trained by minimizing a weighted combination of: 1) A ranking loss on labeled training data, and 2) A distillation loss that encourages ranking unlabeled items similar to the teacher's top-K predictions. The weighting scheme dynamically adjusts the importance of each document based on position and ranking discrepancy between teacher and student. \\
\midrule

\multicolumn{2}{l}{\textbf{Case 27}} \\
\textit{Problem Structure:} & \textit{Problem Structure:} \\
The problem is structured as a hierarchical knowledge transfer challenge where complex user-item relationships need to be decomposed into specialized knowledge domains while maintaining the ability to reconstruct comprehensive recommendations through expert collaboration. & The problem is framed as needing to effectively transfer the vast latent knowledge from a teacher model's representation space to a student model with limited capacity. This requires both summarizing the knowledge in a way that preserves important relationships and patterns, while also ensuring different aspects of the knowledge are distilled appropriately based on their relationships. \\
\textit{Design Rationale:} & \textit{Design Rationale:} \\
Using specialized expert networks allows for focused knowledge transfer in specific relationship domains while keeping the overall student model capacity manageable The hierarchical structure with dynamic routing enables efficient knowledge organization and flexible adaptation to different types of user-item relationships Incorporating probabilistic sampling and adaptive compression ensures that the most important knowledge is prioritized during transfer & Using multiple experts rather than a single one prevents the mixing of weakly correlated information that could dilute or interfere with the knowledge transfer The selection mechanism allows each expert to specialize in transferring knowledge for strongly correlated entities, leading to more focused and effective knowledge transfer Basing expert selection on correlations in the teacher's representation space allows the system to naturally decompose the knowledge along meaningful boundaries \\
\textit{Universal Principle:} & \textit{Universal Principle:} \\
Complex knowledge can be effectively transferred by decomposing it into specialized domains while maintaining cross-domain relationships through hierarchical organization and dynamic routing. & Complex knowledge transfer between systems of different capacities can be made more effective by decomposing the knowledge into coherent subsets and using specialized transfer channels for each subset. \\
\textit{Key Mechanism:} & \textit{Key Mechanism:} \\
The system employs multiple specialized expert networks, each trained on a specific cluster of user-item relationship patterns identified through probabilistic rank-aware sampling. A gating mechanism dynamically routes inputs to relevant experts based on relationship type classification. Knowledge is transferred using a temperature-controlled distillation process with boundary-aware loss functions to maintain clear knowledge separation. The experts' outputs are combined through an adaptive weighting scheme that considers both local and global importance. & Multiple expert neural networks are trained to reconstruct the teacher's internal representations from the student's representations. A selection network determines which expert should handle each piece of knowledge based on correlations in the teacher's representation space. The selection uses Gumbel-Softmax for differentiable expert choice during training. Each expert specializes in transferring knowledge for a particular subset of strongly correlated entities. \\
\midrule
\multicolumn{2}{l}{\textbf{Case 28}} \\
\textit{Problem Structure:} & \textit{Problem Structure:} \\
The challenge is structured as a multi-level knowledge transfer problem where unbiased patterns need to be preserved while handling multiple types of bias. The solution uses a hierarchical approach combining general recommendation knowledge with specialized bias handling. & The problem is structured as a knowledge transfer task where a small amount of unbiased (uniform) data can be used to guide and improve the learning on a larger biased dataset through various forms of knowledge distillation. The framework divides the knowledge transfer into four categories based on what type of information is being distilled: labels, features, samples, or model structure. \\
\textit{Design Rationale:} & \textit{Design Rationale:} \\
A teacher-student architecture allows effective transfer of unbiased patterns from the small uniform dataset while maintaining scalability Specialist models for different bias types enable targeted bias mitigation while preserving general recommendation capability Temperature scaling in knowledge distillation helps control the granularity of transferred patterns & Breaking down knowledge transfer into four distinct categories (labels, features, samples, structure) provides multiple complementary ways to leverage the uniform data, increasing the framework's flexibility and effectiveness Using knowledge distillation as the transfer mechanism allows for soft transfer of information rather than hard constraints, making the framework more robust and adaptable The modular design allows different distillation approaches to be combined or used independently based on the specific needs and constraints of the application \\
\textit{Universal Principle:} & \textit{Universal Principle:} \\
Complex system biases can be effectively addressed through a combination of general knowledge transfer and specialized handling of specific bias patterns & Limited but high-quality data can be systematically decomposed into different aspects (labels, features, samples, structure) to guide and improve learning on larger but imperfect datasets. \\
\textit{Key Mechanism:} & \textit{Key Mechanism:} \\
Train a teacher model on uniform data, then use temperature-scaled soft targets to train both a general student model and specialist models for each bias type. The final system combines predictions using: score = $\alpha *general_pred + \sum ( \beta i * specialist_predi) $, where $\alpha$ and $\beta$i are dynamically adjusted based on detected bias types. & The framework enables uniform data modeling through four approaches: 1) Label-based distillation using imputed labels as guidance 2) Feature-based distillation to filter representative causal features 3) Sample-based distillation for mutual learning between uniform and non-uniform data 4) Model structure-based distillation using embedded representations. \\
\midrule


\multicolumn{2}{l}{\textbf{Case 31}} \\
\textit{Problem Structure:} & \textit{Problem Structure:} \\
The problem is structured as a need to learn hierarchical representations from large-scale bipartite e-commerce graphs while preserving non-linear interactions. The solution approaches this by combining separate processing streams for users and items with hierarchical coarsening that maintains the bipartite structure. & The problem is structured as a multi-level representation learning task on bipartite graphs, where each level captures increasingly abstract patterns of user-item interactions through alternating GNN processing and clustering steps. \\
\textit{Design Rationale:} & \textit{Design Rationale:} \\
Separating user and item processing streams allows for specialized representation learning while maintaining the bipartite nature of e-commerce interactions Using soft clustering with differentiable pooling enables end-to-end training while creating interpretable hierarchical structures Incorporating both interaction-based and content-based similarities ensures rich representations that capture both behavioral and semantic relationships & The alternating GNN-clustering approach enables capturing both detailed local interactions and broader community-level patterns while maintaining computational efficiency Using deterministic clustering rather than learned pooling makes the approach scalable to very large graphs The hierarchical structure allows the model to capture patterns at multiple scales, from individual preferences to community-level trends \\
\textit{Universal Principle:} & \textit{Universal Principle:} \\
Complex hierarchical structures in large-scale networks can be effectively learned by combining parallel specialized processing streams with soft clustering mechanisms that preserve domain-specific constraints. & Iterative abstraction through alternating representation learning and clustering can efficiently capture hierarchical patterns in large-scale relational data. \\
\textit{Key Mechanism:} & \textit{Key Mechanism:} \\
The solution employs two parallel GNNs at each hierarchical level - one for embedding generation and another for clustering. The clustering GNN produces separate assignment matrices for users and items, maintaining the bipartite structure. The model uses differentiable pooling to create coarsened graphs while preserving non-linear interactions through skip connections and attention mechanisms. & HiGNN operates by: 1) Using GNN modules to learn representations of nodes at each level, 2) Applying clustering to group similar nodes and create a coarsened graph, 3) Repeating this process multiple times to build a hierarchical structure. Each level's GNN captures increasingly abstract patterns while the clustering step enables efficient scaling by reducing graph size. \\
\midrule

\multicolumn{2}{l}{\textbf{Case 33}} \\
\textit{Problem Structure:} & \textit{Problem Structure:} \\
The problem is reframed from a global embedding learning task to a local pattern recognition task in a bipartite graph. Instead of learning user/item-specific embeddings, the focus is on identifying and leveraging generalizable local graph patterns that can predict interactions for any user-item pair, including unseen ones. & The problem is reframed as learning local graph patterns around user-item pairs that are predictive of ratings, rather than learning global latent embeddings. This structure treats rating prediction as a graph-level regression task on enclosing subgraphs, allowing generalization to unseen nodes through structural patterns. \\
\textit{Design Rationale:} & \textit{Design Rationale:} \\
Using local graph patterns instead of global embeddings enables natural generalization to unseen users/items since the patterns learned are independent of specific node identities Graph neural networks with message passing provide a powerful framework for learning and encoding these local patterns while maintaining inductive capabilities The combination of structural regularization techniques (node dropout, edge masking) ensures robust pattern learning that can transfer to new scenarios & Using local graph patterns instead of global embeddings enables generalization to unseen entities through structural similarity, making the model inductive without requiring side information The R-GCN message passing architecture allows learning rich interaction patterns between different rating types while preserving their ordinal relationships through regularization Focusing pooling on only target nodes ensures prediction is based on their specific roles in the local graph pattern rather than treating all nodes equally \\
\textit{Universal Principle:} & \textit{Universal Principle:} \\
Complex global behaviors can be understood and predicted through the analysis of local patterns and structures, enabling generalization to unseen entities without requiring specific information about them. & Local interaction patterns between entities and their immediate neighbors contain sufficient information to characterize their direct relationships, enabling generalization through structural similarity rather than entity-specific features. \\
\textit{Key Mechanism:} & \textit{Key Mechanism:} \\
For each user-item pair, extract a k-hop local subgraph. Process this subgraph through a GNN with structural message passing to encode local patterns. Apply node/edge dropout during training to ensure robustness. Use the encoded patterns to predict interaction values through a decoder network. For new users/items, apply the same process to their local neighborhoods. & For each user-item pair, IGMC: 1) Extracts an h-hop enclosing subgraph around the pair from the rating matrix viewed as a bipartite graph, 2) Labels nodes based on their type and distance to target pair, 3) Uses a graph neural network with R-GCN layers to process the subgraph and predict the rating, focusing only on the target user and item nodes in the final pooling step. \\
\midrule
\multicolumn{2}{l}{\textbf{Case 34}} \\
\textit{Problem Structure:} & \textit{Problem Structure:} \\
The problem is structured as a two-stage process: first capturing rich contextual information for each word through bi-directional recurrent processing, then identifying the most salient features through convolution and max-pooling operations. This allows the model to maintain both broad context and focused feature selection. & The challenge is framed as creating a hybrid architecture that combines the sequential processing capabilities of recurrent networks with the feature selection strengths of convolutional networks. This requires balancing local and global context while maintaining computational efficiency. \\
\textit{Design Rationale:} & \textit{Design Rationale:} \\
The bidirectional RNN component ensures comprehensive context capture from both past and future words, while being more computationally efficient than processing large fixed windows The hierarchical convolutional structure with multiple kernel sizes enables the model to capture features at different granularities while maintaining efficiency through shared parameters The integration of pre-trained word embeddings provides rich linguistic features that can be fine-tuned for specific tasks & The bi-directional recurrent structure allows capturing long-range dependencies without the limitations of fixed window sizes, reducing noise compared to traditional window-based approaches The max-pooling layer automatically identifies the most discriminative features in the sequence, eliminating the need for manual feature engineering while maintaining computational efficiency \\
\textit{Universal Principle:} & \textit{Universal Principle:} \\
Complex sequential information can be effectively processed by combining broad context capture with focused feature selection at multiple scales & Complex sequential patterns can be effectively processed by combining continuous context tracking with selective feature extraction mechanisms. \\
\textit{Key Mechanism:} & \textit{Key Mechanism:} \\
The model first processes word embeddings through a bidirectional RNN to create context-aware representations. These representations are then fed into multiple parallel convolutional layers with different kernel sizes (e.g., 3, 4, 5 words). Max pooling is applied to each feature map to select the most salient features, which are concatenated for final classification. & The model uses a bi-directional recurrent structure to capture contextual information, representing each word as a combination of its word embedding and both left and right contexts. These word representations are then processed through a max-pooling layer to automatically select the most important features for classification. The mechanism can be formalized as: Final document representation is obtained through max-pooling over transformed word representations \\
\midrule
\multicolumn{2}{l}{\textbf{Case 35}} \\
\textit{Problem Structure:} & \textit{Problem Structure:} \\
The problem is structured as an iterative refinement process where lexical preferences are treated as protected elements that can be repositioned but not removed, while the surrounding context is dynamically refined to maintain fluency and coherence. & The problem is framed as needing a parallel editing operation that can directly move tokens to their correct positions while handling lexical choice and reordering decisions simultaneously, rather than decomposing them into sequences of more basic operations. \\
\textit{Design Rationale:} & \textit{Design Rationale:} \\
Separating lexical choice from positioning allows for more efficient incorporation of preferences while maintaining translation quality Using parallel refinement operations enables faster decoding compared to traditional autoregressive approaches Protecting user preferences while allowing repositioning strikes an optimal balance between respecting user input and maintaining translation fluency & Making repositioning decisions independent and parallel enables faster decoding compared to sequential edits Combining deletion and repositioning into a single operation reduces the number of steps needed while maintaining trainability Constraining repositioning to use only input tokens preserves the ability to use efficient oracles during training \\
\textit{Universal Principle:} & \textit{Universal Principle:} \\
Complex constraints can be effectively handled by protecting core elements while allowing flexible refinement of their context, enabling both constraint satisfaction and overall quality optimization. & Complex sequential transformations can be made more efficient by identifying atomic parallel operations that handle multiple aspects simultaneously while preserving properties needed for optimization. \\
\textit{Key Mechanism:} & \textit{Key Mechanism:} \\
The solution uses a modified Levenshtein Transformer that initializes translation with user preferences, protects them from deletion, and applies parallel insertion/deletion operations to refine the surrounding context. It employs confidence-based masking to identify and improve low-confidence regions while preserving user-specified terms. & For each position i in the input sequence, independently predict whether to delete the token (r=0) or replace it with another input token yr (r>0). The reposition operation maintains sequence boundaries by constraining 1. Multiple repositions can be applied in parallel during decoding. \\
\midrule
\multicolumn{2}{l}{\textbf{Case 36}} \\
\textit{Problem Structure:} & \textit{Problem Structure:} \\
The problem is structured as a hierarchical feature integration challenge where features at different scales need to be effectively combined while adaptively weighting their importance. The solution focuses on creating dynamic pathways between encoder and decoder layers while maintaining computational efficiency. & The problem is framed as needing to combine and realign semantic features from multiple scales while adaptively weighting channel importance, bridging the semantic gap between encoder and decoder layers. \\
\textit{Design Rationale:} & \textit{Design Rationale:} \\
Full-scale skip connections with adaptive channel attention enable comprehensive information flow while intelligently filtering irrelevant features Asymmetric convolution blocks provide efficient multi-scale feature extraction by decomposing spatial relationships Progressive channel fusion with learned importance weights ensures optimal feature combination while managing computational complexity & Direct connections between all scales allow the network to leverage complementary information from different processing depths without being constrained by the hierarchical structure Channel attention helps focus on the most relevant features from each scale by learning to weight channels based on their informativeness Using both max and average pooling in the attention mechanism captures different aspects of channel importance \\
\textit{Universal Principle:} & \textit{Universal Principle:} \\
Complex hierarchical information can be effectively integrated by creating adaptive pathways that dynamically weight the importance of different information streams based on context & Multi-scale information can be more effectively combined by allowing direct interaction between all scales while using attention mechanisms to determine relative importance. \\
\textit{Key Mechanism:} & \textit{Key Mechanism:} \\
Implements full-scale skip connections between all encoder-decoder layers, with each connection passing through an adaptive channel attention module. Features are processed using asymmetric convolution blocks (square, horizontal, vertical kernels) at each scale. A progressive channel fusion mechanism combines features through learned weights and channel shuffling operations. The channel attention mechanism uses both global average and max pooling followed by shared MLPs to generate attention weights. & Features from encoder layers at all depths are connected to each decoder layer through skip connections. A channel attention block uses both max and average pooling followed by MLP layers to generate channel-wise weights that adaptively recalibrate feature channels before combination. \\
\midrule
\multicolumn{2}{l}{\textbf{Case 37}} \\
\textit{Problem Structure:} & \textit{Problem Structure:} \\
The problem is structured as a multi-level encoding challenge where both word-level and sentence-level importance need to be captured simultaneously while maintaining the hierarchical nature of documents. The solution recognizes that document understanding requires both local (word) and global (sentence) context processing. & The problem is framed as a hierarchical document representation task where importance is determined at two levels - words within sentences and sentences within documents - with each level requiring its own attention mechanism to identify and weight relevant content based on context. \\
\textit{Design Rationale:} & \textit{Design Rationale:} \\
Combining bidirectional GRU with dual-level attention mechanisms allows for capturing context in both directions while selectively focusing on important elements at each level Using a hierarchical structure mirrors the natural organization of documents (words→sentences→document), making the model more interpretable and effective Incorporating adaptive pooling mechanisms helps handle variable-length inputs while preserving important information at each level & The hierarchical structure mirrors the natural organization of documents, allowing the model to capture relationships at appropriate levels of granularity Using attention at both word and sentence levels enables the model to dynamically focus on relevant content, rather than treating all content equally The context vectors at each level provide a learned mechanism for determining importance that can adapt to different types of documents and classification tasks \\
\textit{Universal Principle:} & \textit{Universal Principle:} \\
Complex hierarchical structures can be effectively processed by using multi-level attention mechanisms that selectively focus on important elements while maintaining contextual relationships between different levels of the hierarchy. & Complex hierarchical structures are best understood by processing them at multiple levels of granularity while dynamically determining the importance of components at each level. \\
\textit{Key Mechanism:} & \textit{Key Mechanism:} \\
The model processes text in two stages: First, words are encoded using bidirectional GRU with word-level attention to create sentence representations. Then, sentences are encoded using another bidirectional GRU with sentence-level attention to create the final document representation. Both attention mechanisms compute importance weights based on learned context vectors. & The model uses a two-level architecture: 1) Word-level: A bidirectional GRU processes words to get contextual word annotations, which are weighted by attention scores computed using a trainable context vector to form sentence vectors. 2) Sentence-level: Another bidirectional GRU processes sentence vectors, which are weighted by sentence-level attention scores to form the final document representation. The attention scores at each level are computed as: $\alpha$ = $softmax(u^T tanh(Wh + b))$, where u is a trainable context vector. \\
\midrule

\multicolumn{2}{l}{\textbf{Case 39}} \\
\textit{Problem Structure:} & \textit{Problem Structure:} \\
The problem is structured as a dual-stream learning challenge where entity recognition and contextual understanding are separated but coordinated. This approach allows for effective knowledge transfer from pre-trained models while handling noisy distant supervision by maintaining distinct but complementary learning paths. & The problem is structured as a two-stage learning process where first, a pre-trained language model adapts to the NER task using distant labels, and second, a self-training approach refines the model using generated pseudo-labels. This assumes that pre-trained models contain transferable knowledge that can help overcome noisy/incomplete supervision. \\
\textit{Design Rationale:} & \textit{Design Rationale:} \\
The separation of entity and context streams allows for more robust handling of noisy labels since the context stream can maintain stable linguistic knowledge even when entity labels are uncertain Using entity-focused masking during pre-training helps the model develop strong entity representations before encountering noisy distant labels The two-stream architecture enables selective updating during self-training, allowing the model to preserve valuable pre-trained knowledge while refining entity recognition capabilities & Using pre-trained language models provides rich semantic knowledge that helps overcome noisy/incomplete distant labels by better understanding context and relationships between words Two-stage approach allows first leveraging transferred knowledge, then refining it through self-training, providing complementary benefits at each stage Early stopping and confidence-based selection help prevent overfitting to noise while allowing the model to learn from high-quality predictions \\
\textit{Universal Principle:} & \textit{Universal Principle:} \\
Complex learning tasks can be made more robust by separating and independently managing different aspects of the problem while maintaining coordination between them. & Complex supervised learning tasks can be improved by first transferring knowledge from pre-trained models, then progressively refining that knowledge through iterative self-training. \\
\textit{Key Mechanism:} & \textit{Key Mechanism:} \\
The solution implements two parallel attention streams: an entity identification stream and a context understanding stream. The entity stream focuses on boundary detection and type classification using entity-focused masking and selective updating, while the context stream maintains stable linguistic knowledge. During training, the streams share parameters but use different attention masks, with updates primarily affecting the entity stream during refinement phases. & First stage adapts BERT/RoBERTa to NER using distant labels with early stopping to avoid overfitting. Second stage uses self-training where a teacher model generates pseudo-labels for student model training. Student becomes teacher for next iteration. High-confidence soft labels are used to improve training quality. \\
\midrule



\multicolumn{2}{l}{\textbf{Case 43}} \\
\textit{Problem Structure:} & \textit{Problem Structure:} \\
The problem is framed as a graph representation challenge where characters and potential words are nodes in a dynamic graph structure, allowing the model to learn and select relevant information without committing to rigid word segmentation. & The problem is structured as encoding both character sequences and potential word sequences simultaneously in a neural network, where word candidates come from a lexicon rather than segmentation. The model needs to dynamically determine which word combinations are most useful for NER tagging. \\
\textit{Design Rationale:} & \textit{Design Rationale:} \\
Using a graph structure with both character and word nodes allows explicit modeling of relationships while maintaining flexibility in word boundary decisions Graph attention mechanisms enable dynamic weighting of character vs word information based on context The combination of CNN for character features and GNN for structure provides rich representations at multiple levels & Using a lattice structure allows incorporating all potential word matches from a lexicon while maintaining character-level processing, providing more flexibility than fixed word segmentation Gated information flow from word cells to character cells enables the model to automatically learn which word combinations are most useful for the target task Building on character-level LSTM-CRF provides robust sequential modeling while allowing integration of word information \\
\textit{Universal Principle:} & \textit{Universal Principle:} \\
Complex hierarchical relationships can be effectively modeled through graph structures that allow dynamic information flow between different levels of abstraction & Complex sequences can be better understood by considering both their atomic units and potential higher-level patterns simultaneously, with dynamic selection of the most relevant patterns at each point. \\
\textit{Key Mechanism:} & \textit{Key Mechanism:} \\
The model constructs a graph where characters and potential words are nodes, connected by edges representing possible word formations. A character-level CNN extracts features from individual characters, while a Graph Attention Network processes the entire structure to enable dynamic information exchange. The final representation for each character position combines local CNN features with attended graph-level information, feeding into a CRF layer for NER prediction. & A lattice-structured LSTM that takes both characters and matched lexicon words as input. For each character, there are multiple paths leading to it through different matched words. Gated cells control information flow from these paths to construct character representations that incorporate relevant word information. The model uses:   - Character cells c[c]j for recording sequential information   - Word cells c[w]b,e for word information    - Gates for controlling information flow from word cells to character cells   - Character hidden vectors h[c]j for the final CRF layer \\
\midrule
\multicolumn{2}{l}{\textbf{Case 44}} \\
\textit{Problem Structure:} & \textit{Problem Structure:} \\
The problem is structured as a dual-stream processing challenge where character-level and word-level information need to be processed simultaneously while resolving boundary conflicts through iterative refinement using semantic context. & The problem is framed as a need for parallel processing of characters and lexicons while having a feedback mechanism to refine word selection based on high-level semantic understanding. The structure assumes that word conflicts can be resolved by feeding back contextual information from higher layers to guide lexicon attention. \\
\textit{Design Rationale:} & \textit{Design Rationale:} \\
Using parallel dilated CNN channels allows efficient processing of both character and word information while capturing broad context through exponentially increasing dilation rates The iterative refinement mechanism with feedback layers enables the resolution of boundary conflicts by incorporating higher-level semantic information The emphasis vector mechanism allows dynamic adjustment of feature importance based on initial predictions, helping resolve ambiguities in subsequent iterations & Using CNNs with parallel processing enables much faster computation compared to sequential RNN-based approaches, while still capturing multi-scale pattern information through stacked layers The feedback mechanism allows high-level semantic understanding to influence lower-level pattern matching decisions, helping resolve ambiguities that can't be solved with only local context Multi-scale feature attention adaptively selects features of different scales, allowing the model to focus on the most relevant pattern granularity for each position \\
\textit{Universal Principle:} & \textit{Universal Principle:} \\
Complex sequential processing tasks can be effectively handled through parallel processing streams combined with iterative refinement mechanisms that leverage higher-level contextual information to resolve local ambiguities. & Global context can be used to refine local pattern matching through feedback loops, enabling more accurate disambiguation of competing interpretations. \\
\textit{Key Mechanism:} & \textit{Key Mechanism:} \\
The model employs two parallel dilated CNN channels (character and word) with exponentially increasing dilation rates (1,2,4,8). After initial processing, an emphasis layer generates vectors based on preliminary predictions to modulate feature maps in both channels. This process iterates multiple times, with each iteration refining boundary decisions using feedback from higher-level semantic understanding. & The model uses CNNs to process characters and potential matching words in parallel, with an attention mechanism to incorporate lexicon information. A rethinking mechanism feeds back high-level features from the top layer to refine the lexicon attention weights at each layer using:$ i2 = \alpha (Wi*Cml + Ui*wml + ViXmL + bi*)$, where XmL represents high-level features used to readjust attention weights. \\
\midrule
\multicolumn{2}{l}{\textbf{Case 45}} \\
\textit{Problem Structure:} & \textit{Problem Structure:} \\
The problem is structured as a multi-layered challenge requiring both specialized storage/processing for different data types and intelligent cross-domain connection discovery. This approach recognizes that effective hypothesis generation requires both preserving the unique characteristics of different biomedical data types while enabling novel cross-domain discoveries. & The problem is framed as creating a multi-modal network that efficiently connects biomedical concepts through both direct (known) and indirect (implicit) relationships, using a combination of structured databases and unstructured text analysis to enable discovery of non-obvious connections. \\
\textit{Design Rationale:} & \textit{Design Rationale:} \\
The multi-dimensional graph structure with specialized processing units optimizes handling of different data types while maintaining their interconnectedness The two-phase analysis approach (statistical pattern mining followed by cross-domain integration) ensures both statistical rigor and novel discovery potential The pipeline processing system with entity bundling enables efficient processing of related entities while maintaining scalability & Using multiple data types and connection methods ensures both precision from structured data and discovery potential from unstructured text analysis Weighted connections allow the system to favor more reliable or relevant paths when generating hypotheses The multi-layer approach separates different types of relationships while allowing interaction between layers through cross-connections \\
\textit{Universal Principle:} & \textit{Universal Principle:} \\
Complex systems can be effectively analyzed by combining specialized processing of individual components with intelligent integration mechanisms that preserve component-specific characteristics while enabling cross-component discoveries. & Complex relationships between concepts can be discovered by analyzing both explicit connections in structured data and implicit connections derived from unstructured information. \\
\textit{Key Mechanism:} & \textit{Key Mechanism:} \\
The solution implements a multi-dimensional knowledge graph where each dimension represents a different biomedical data type, managed by specialized processing units. Statistical pattern mining identifies significant relationships within each dimension, which then serve as constraints for cross-dimensional analysis. A pipeline processing system with entity bundling enables efficient data retrieval and analysis across dimensions. & The system constructs a weighted multi-layer network where: 1) Abstract nodes are connected based on semantic similarity using FastText vectors and FLANN, 2) Keyword nodes are connected using UMLS relationships, and 3) Cross-layer connections are created using tf-idf scores. Shortest path queries between concepts identify potential relationship chains for hypothesis generation. \\
\midrule
\multicolumn{2}{l}{\textbf{Case 46}} \\
\textit{Problem Structure:} & \textit{Problem Structure:} \\
The problem is structured as a multi-level optimization challenge where motif instances need to be evaluated and selected based on both their local semantic relevance and their significance at different taxonomy levels. This requires balancing local context with global hierarchical importance. & The problem is framed as a need to identify and filter the most relevant motif instances for a given taxonomy node based on its granularity and semantics, treating motif selection as a dynamic process that adapts to the current clustering task rather than a static one-time selection. \\
\textit{Design Rationale:} & \textit{Design Rationale:} \\
Combining dynamic motif conductance with hierarchical significance testing provides a robust way to evaluate motif instances at different granularity levels while maintaining computational efficiency Incorporating semantic context through meta-paths ensures that selected motif instances are not just structurally significant but also semantically meaningful Using adaptive thresholding allows the framework to adjust its selectivity based on the taxonomy level, ensuring appropriate granularity of information at each level & Using anchor terms from initial clusters provides a reliable foundation for evaluating motif instances, as these terms best represent the desired clustering structure The importance and concentration metrics complement each other - importance ensures influential evidence while concentration ensures discriminative power The instance-level selection allows the system to be selective even within generally useful motif patterns, providing finer-grained control than pattern-level selection \\
\textit{Universal Principle:} & \textit{Universal Principle:} \\
Complex hierarchical systems can be effectively analyzed by combining local contextual information with global structural significance, using adaptive evaluation metrics that vary with the hierarchical level. & The relevance and usefulness of relationship evidence depends on the granularity and context of the organizational task, requiring dynamic selection rather than static application. \\
\textit{Key Mechanism:} & \textit{Key Mechanism:} \\
The framework operates by: 1) Computing a Dynamic Motif Instance Conductance (DMIC) score for each instance that considers both structural and semantic properties, 2) Generating level-specific null models to assess significance, 3) Using adaptive thresholds based on taxonomy level to select instances, and 4) Applying the formula: Score(m,l) = $\alpha$(l)*DMIC(m) + (1-$\alpha$(l))*SemanticRelevance(m), where $\alpha$(l) is a level-dependent weighting factor. & At each taxonomy node: 1. Get initial term clusters from text-based embedding 2. Select anchor terms from each cluster through comparative analysis  3. Score motif instances based on:    - Importance: mean importance score across clusters from authority ranking    - Concentration: normalized entropy of importance scores across clusters    Final score = Sqrt(importance × concentration) 4. Select top K\% motif instances based on scores \\
\midrule
\multicolumn{2}{l}{\textbf{Case 47}} \\
\textit{Problem Structure:} & \textit{Problem Structure:} \\
The problem is structured as a parameter efficiency challenge in deep CNN architectures for super-resolution, where the goal is to achieve high performance with minimal parameters while maintaining stability at extreme depths. The solution approaches this through a hierarchical structure that combines recursive processing with multi-level residual learning. & The problem is structured as building deep networks through recursive blocks that combine local and global residual learning. Each recursive block contains multiple residual units sharing weights, while different recursive blocks use different weights. This structure allows increasing depth without proportionally increasing parameters. \\
\textit{Design Rationale:} & \textit{Design Rationale:} \\
Recursive weight sharing dramatically reduces parameter count while maintaining effective network depth, allowing for deep feature extraction without linear parameter growth Multi-level residual learning (both global and local) ensures stable gradient flow and prevents performance degradation in very deep architectures Hierarchical structure with bottleneck designs enables efficient processing of both local and global features while further reducing parameters & Recursive weight sharing within blocks reduces parameters while still allowing network depth to increase, making very deep networks practical with limited memory/storage Multi-path structure with both local and global residual connections helps gradient flow during training and preserves information throughout the deep network Separation of recursive blocks with different weights provides flexibility to learn diverse features while keeping the model compact through internal weight sharing \\
\textit{Universal Principle:} & \textit{Universal Principle:} \\
Complex system performance can be maintained or improved while reducing resource requirements by combining hierarchical structures with recursive processing and residual connections & Complex transformations can be learned efficiently by recursively applying shared components while maintaining direct paths for information flow at multiple levels. \\
\textit{Key Mechanism:} & \textit{Key Mechanism:} \\
The network consists of hierarchical stages, each containing recursive blocks that share weights. Each recursive block uses a bottleneck structure (1x1 -> 3x3 -> 1x1 convolutions) and implements local residual learning. The stages are connected through skip connections, and the entire network operates within a global residual learning framework. The final output is produced by combining intermediate outputs from different recursion depths using learned weights. & DRRN consists of B recursive blocks, each containing U residual units that share weights. Each residual unit has dual branches - a residual branch with two conv layers and an identity branch connecting to the input. Global residual learning is achieved by adding the final output to the input image. The mathematical formulation is: y = fRec(RB(RB-1(...(R1(x))...))) + x, where Rb represents the b-th recursive block function. \\
\midrule



\multicolumn{2}{l}{\textbf{Case 51}} \\
\textit{Problem Structure:} & \textit{Problem Structure:} \\
The problem is structured as a multi-level learning challenge where language-independent acoustic features need to be captured in shared layers while maintaining language-specific characteristics through separate output layers. This approach eliminates the need for frame-level alignments while efficiently handling multiple languages with varying resource availability. & The problem is structured as learning a shared representation across languages through sequence modeling, where the lower layers learn language-independent features while language-specific knowledge is captured in the final projection layer. This eliminates the need for explicit alignments while enabling cross-lingual transfer. \\
\textit{Design Rationale:} & \textit{Design Rationale:} \\
The bottleneck architecture forces the model to learn a compressed, language-independent representation that can generalize across languages, making it particularly effective for low-resource scenarios. The combination of shared encoding layers with language-specific softmax layers allows the model to balance universal acoustic patterns with language-specific characteristics. The interval-based softmax mechanism enables efficient computation and training by only activating relevant language outputs, reducing computational overhead and preventing interference between languages. & Sharing lower layers across languages enables learning more robust acoustic features by leveraging data from multiple languages, providing better regularization Using CTC loss eliminates the need for explicit alignments, greatly simplifying the training process compared to traditional HMM-based approaches Language-specific linear projections provide a lightweight way to adapt the shared representations to different phone sets while keeping most parameters shared \\
\textit{Universal Principle:} & \textit{Universal Principle:} \\
Complex multi-domain problems can be solved by creating a compressed shared representation of common features while maintaining domain-specific outputs, allowing efficient knowledge transfer from resource-rich to resource-poor domains. & Complex patterns shared across multiple domains can be learned through a common representation layer while maintaining domain-specific outputs through simple linear transformations. \\
\textit{Key Mechanism:} & \textit{Key Mechanism:} \\
The model uses a deep neural network with shared encoding layers, including a bottleneck layer, followed by language-specific softmax layers. Training occurs simultaneously across all languages using sequence-based methods (CTC/attention). The interval-based softmax is applied only to relevant language outputs during forward and backward passes. & The model uses stacked Bidirectional LSTM layers shared across languages for encoding acoustic features, followed by language-specific linear projections to phone sets. Training uses CTC loss which automatically learns alignments between input and output sequences by introducing a blank symbol. The shared layers learn language-independent features while the projections handle language-specific mappings. \\
\midrule
\multicolumn{2}{l}{\textbf{Case 52}} \\
\textit{Problem Structure:} & \textit{Problem Structure:} \\
The problem is structured as an iterative self-improvement cycle where limited initial data is maximally leveraged through augmentation, followed by controlled expansion of the training dataset through selective incorporation of automatically transcribed data. & The problem is framed as a bootstrapping process where an initial model trained on limited labeled data is used to automatically transcribe unlabeled data, with quality control through confidence scoring to add only reliable transcriptions for model retraining. \\
\textit{Design Rationale:} & \textit{Design Rationale:} \\
Speed perturbation provides a reliable way to triple the initial training data without introducing artificial artifacts, creating natural variations in speech rate and pitch Using an ensemble-based confidence scoring system with Leaky HMM regularization helps prevent the reinforcement of early errors while maintaining model flexibility The adaptive threshold mechanism ensures that data quality standards increase as the model improves, preventing degradation of the training set & Using confidence scores provides an automated way to assess transcription quality without requiring manual verification, making the process scalable Iterative retraining allows the model to gradually improve and generate better quality transcriptions in subsequent passes Setting the threshold at the average confidence creates a balanced approach to data filtering, neither too strict nor too permissive \\
\textit{Universal Principle:} & \textit{Universal Principle:} \\
Limited high-quality data can be effectively expanded through a combination of natural augmentation and carefully controlled self-training, where the system gradually raises its own standards as it improves. & Systems can bootstrap themselves to better performance by using their own outputs as additional training data, provided there is a reliable way to assess output quality. \\
\textit{Key Mechanism:} & \textit{Key Mechanism:} \\
Initially applies speed perturbation (0.9x, 1.0x, 1.1x) to triple the training data. Trains an ensemble of models with Leaky HMM regularization. Uses cross-model agreement and confidence scores to select new training examples. Gradually increases confidence thresholds as performance improves. & 1. Train initial model on small manually transcribed dataset 2. Use model to automatically transcribe untranscribed data 3. Calculate average confidence score across all automatic transcriptions 4. Filter out transcriptions below confidence threshold 5. Add remaining high-confidence transcriptions to training data 6. Retrain model and repeat process iteratively \\
\midrule
\multicolumn{2}{l}{\textbf{Case 53}} \\
\textit{Problem Structure:} & \textit{Problem Structure:} \\
The problem is structured as a dual optimization challenge where the goal is to create a bottleneck architecture that can simultaneously preserve speech recognition capability while actively removing language-specific information through adversarial training. & The problem is framed as a need to ensure shared neural network layers learn language-invariant features while still capturing useful acoustic patterns. This involves creating an adversarial dynamic between the main speech recognition task and a language identification task to force language-independent representations. \\
\textit{Design Rationale:} & \textit{Design Rationale:} \\
The bottleneck architecture naturally constrains the information flow, forcing the model to learn efficient, compressed representations that focus on the most essential features for speech recognition Progressive adversarial training allows the model to first establish good speech recognition capabilities before gradually introducing language-invariance constraints, preventing early convergence to poor solutions The use of reconstruction loss helps ensure that the compressed representations retain acoustically relevant information while the adversarial component removes language-specific features & Gradient reversal forces shared layers to learn features that are useful for the main task but uninformative about unwanted categorical attributes, making the representations more transferable Gradual increase of the adversarial weight allows the model to first learn good task-specific features before focusing on removing biases, providing more stable training The adversarial language classifier provides an explicit training signal for removing language-specific information, rather than relying on implicit regularization \\
\textit{Universal Principle:} & \textit{Universal Principle:} \\
Complex competing objectives can be balanced through progressive constraint introduction and information bottlenecking, allowing systems to learn essential features while discarding unwanted characteristics. & Adversarial training can be used to explicitly remove unwanted biases or attributes from learned representations while preserving task-relevant information. \\
\textit{Key Mechanism:} & \textit{Key Mechanism:} \\
The model consists of an encoder with bottleneck layers feeding into three branches: speech recognition, reconstruction, and language classification (with gradient reversal). Training proceeds in phases, gradually increasing the weight of the adversarial language classification loss while maintaining speech recognition and reconstruction performance. The bottleneck forces compact representations while the gradient reversal ensures language-invariance. & A gradient reversal layer is inserted between shared layers and a language discriminator, causing shared layers to be updated to maximize language classification loss while minimizing speech recognition loss. The loss is: L($\theta$m, $\theta$a, $\theta$s) = LMul($\theta$m, $\theta$s) + $\lambda$LAdv($\theta$a, $\theta$s) where $\lambda$ gradually increases from 0 to 1 during training. \\
\midrule


\multicolumn{2}{l}{\textbf{Case 56}} \\
\textit{Problem Structure:} & \textit{Problem Structure:} \\
The problem is structured as a multi-level semantic information capture challenge in few-shot text classification, where both fine-grained semantic details and high-level class discrimination need to be balanced through a hierarchical attention mechanism. & The problem is framed as a need to create a hierarchical attention mechanism that operates at multiple levels (feature, word, instance) to better capture and weight important information from limited training examples while reducing the impact of noise. \\
\textit{Design Rationale:} & \textit{Design Rationale:} \\
Combining hierarchical attention with prototypical networks allows for capturing semantic information at multiple granularities while maintaining computational efficiency Using instance-level attention with noise detection helps build robust prototypes even with limited and potentially noisy examples The adaptive feature-level attention mechanism enables dynamic adjustment of feature importance based on class relationships, improving discrimination between similar classes & Multiple levels of attention allow the model to capture important information at different semantic levels, making better use of limited training data Hierarchical structure enables the model to progressively refine its focus from low-level features to high-level instance relationships Integration with prototypical networks provides an efficient way to perform classification while maintaining the ability to handle new classes \\
\textit{Universal Principle:} & \textit{Universal Principle:} \\
Complex information processing systems benefit from hierarchical attention mechanisms that can selectively focus on relevant information at multiple levels while maintaining robustness to noise. & Hierarchical processing with attention at multiple levels enables more efficient learning from limited data by selectively focusing on the most relevant information at each level of abstraction. \\
\textit{Key Mechanism:} & \textit{Key Mechanism:} \\
The model employs a three-level attention hierarchy: word-level attention identifies important words, sentence-level attention captures key sentences, and instance-level attention weighs support examples when constructing prototypes. A noise detection module adjusts instance weights based on semantic consistency. The final classification is performed using distance metrics modulated by feature-level attention that emphasizes discriminative features for each class pair. & The model uses three levels of attention mechanisms: 1) Feature level attention that learns feature importance scores for each class, 2) Word level attention that weights important words in instances, and 3) Instance level multi-cross attention that determines instance importance for prototype construction. These are combined with prototypical networks that classify queries based on distances to class prototypes. \\
\midrule
\multicolumn{2}{l}{\textbf{Case 57}} \\
\textit{Problem Structure:} & \textit{Problem Structure:} \\
The problem is structured as a parallel processing challenge where spatial and temporal information need to be processed independently yet combined effectively. The solution uses two complementary pathways to process different aspects of video data, with an adaptive fusion mechanism to optimally combine their outputs. & The problem is framed as a need to combine two complementary approaches for video analysis - one that processes spatial and temporal information sequentially (CNN-RNN) and another that processes them simultaneously (C3D), leveraging their different strengths through late fusion to achieve better overall performance. \\
\textit{Design Rationale:} & \textit{Design Rationale:} \\
The dual-stream architecture allows specialized processing of spatial features through CNN-RNN and spatiotemporal features through C3D, ensuring comprehensive capture of emotion-relevant information Adaptive fusion enables dynamic weighting of pathway contributions based on input characteristics, making the system more robust across different scenarios The use of late fusion preserves the unique characteristics of each pathway while allowing for effective integration of complementary information & The CNN-RNN pathway allows detailed analysis of spatial features before temporal processing, which is particularly important for subtle facial expressions that require fine-grained spatial understanding The C3D pathway enables direct modeling of spatio-temporal patterns without separating spatial and temporal aspects, which can capture integrated motion patterns more effectively Late fusion allows each pathway to develop its own specialized representations before combination, preventing one approach from dominating the other during training \\
\textit{Universal Principle:} & \textit{Universal Principle:} \\
Complex information processing can be enhanced by parallel specialized pathways combined with adaptive integration mechanisms that preserve and leverage the strengths of each pathway. & Complex temporal phenomena can be better understood by combining systems that process temporal relationships both sequentially and holistically. \\
\textit{Key Mechanism:} & \textit{Key Mechanism:} \\
The system processes video through two parallel pathways: 1) CNN-RNN pathway extracts frame-level spatial features and models temporal evolution, 2) C3D pathway processes spatiotemporal features simultaneously. An adaptive fusion module combines outputs using learned weights based on input characteristics. The final emotion classification is produced through a shared classifier layer. & The system uses two main neural network pathways processed in parallel: 1) CNN-RNN pathway: Extracts facial features from individual frames using VGG16 CNN, then processes their temporal relationships using LSTM 2) C3D pathway: Processes fixed-length sequences of frames using 3D convolutions to simultaneously model spatial and temporal patterns The predictions from both pathways are combined through weighted fusion along with audio features for final classification. \\
\midrule

\multicolumn{2}{l}{\textbf{Case 59}} \\
\textit{Problem Structure:} & \textit{Problem Structure:} \\
The challenge is structured as a knowledge transfer problem where a model must learn to leverage abundant data from Chinese while efficiently adapting to limited Tujia data. The solution uses a multi-task architecture that simultaneously learns language-independent features and language-specific characteristics. & The problem is structured as a transfer learning challenge where knowledge from a high-resource language (Chinese) needs to be transferred to benefit a low-resource language (Tujia) model while accounting for both shared and unique linguistic features between the languages. \\
\textit{Design Rationale:} & \textit{Design Rationale:} \\
Combining multi-task learning with selective parameter freezing provides a robust way to prevent overfitting while allowing necessary adaptation to the target language Using shared lower layers for universal acoustic features while maintaining language-specific output layers allows the model to capture both common patterns and unique characteristics The gradual adaptation strategy enables the model to effectively utilize limited data while maintaining performance & Shared hidden layers allow the model to leverage acoustic-phonetic patterns common to both languages, effectively increasing the training data for these shared features. Separate softmax layers maintain the ability to capture language-specific characteristics while benefiting from the shared representations. The use of multiple processing stages (CNN + BiLSTM) allows the model to learn both local and long-range dependencies in speech signals. \\
\textit{Universal Principle:} & \textit{Universal Principle:} \\
Complex systems can effectively transfer knowledge from data-rich to data-poor domains by identifying and preserving universal patterns while selectively adapting specific components. & Complex systems with partial overlap in their underlying structure can share learned representations at intermediate levels while maintaining distinct input and output mappings. \\
\textit{Key Mechanism:} & \textit{Key Mechanism:} \\
The model uses a shared CNN/LSTM architecture with two output branches - one for phoneme recognition and another for language identification. It employs selective parameter freezing where lower layers capturing universal acoustic features are frozen while higher layers are fine-tuned. The loss function combines both tasks: L = $\alpha$Lp(phoneme) + (1-$\alpha$)Li(language), with gradual adaptation as more target language data becomes available. & The model uses CNN and BiLSTM layers to extract acoustic features, with hidden layers shared between Chinese and Tujia language processing paths while maintaining separate softmax output layers with weights {wt, ws, wc} corresponding to Tujia-specific, Chinese-specific, and shared IPA labels. \\
\midrule
\multicolumn{2}{l}{\textbf{Case 60}} \\
\textit{Problem Structure:} & \textit{Problem Structure:} \\
The problem is framed as a need to balance between learning general phonetic features and specific contextual information in low-resource environments. The solution structures this as a hierarchical learning problem where simpler tasks (monophone recognition) provide a foundation for more complex tasks (triphone recognition). & The problem is structured as a need to jointly learn both context-dependent (triphone) and context-independent (monophone) representations through multitask learning, where the two tasks share parameters but maintain separate output layers. This allows the model to leverage the complementary strengths of both representations while regularizing each other. \\
\textit{Design Rationale:} & \textit{Design Rationale:} \\
Progressive training allows the model to first master basic phone discrimination before tackling more complex contextual relationships, preventing overfitting in low-resource scenarios Shared lower layers enable efficient parameter usage and knowledge transfer between tasks while separate output layers maintain task specificity Dynamic weighting between tasks allows the model to adapt to data availability and quality & Using shared hidden layers allows the model to learn common acoustic-phonetic features useful for both tasks while maintaining task-specific output layers The monophone task acts as a regularizer by forcing the model to maintain core phonetic distinctions rather than overfitting to contextual variations \\
\textit{Universal Principle:} & \textit{Universal Principle:} \\
Complex learning tasks can be decomposed into hierarchical subtasks with shared fundamentals, allowing progressive mastery while maintaining efficiency in resource-constrained environments. & Jointly learning multiple levels of abstraction of the same concept can provide implicit regularization and improve model generalization, especially with limited training data. \\
\textit{Key Mechanism:} & \textit{Key Mechanism:} \\
A deep neural network with shared lower layers and separate output layers for monophone and triphone predictions is trained progressively. Training begins with monophone targets using annealed gradient noise, gradually introducing triphone targets as training progresses. The importance of each task is controlled by an adaptive weighting mechanism based on data availability and model performance. & The model learns two tasks in parallel using shared hidden layers but separate output layers - one for monophone prediction and one for triphone prediction. The loss function combines both objectives with a weighting parameter $\alpha$: Loss = -$\sum$(log p(LB[Tri]|x;$\theta$)) - $\alpha$*$\sum$(log p(LB[Mono]|x;$\theta$)) \\
\midrule

\multicolumn{2}{l}{\textbf{Case 62}} \\
\textit{Problem Structure:} & \textit{Problem Structure:} \\
The problem is structured as a hierarchical feature learning challenge where both diverse feature representations and channel interdependencies need to be captured at multiple levels. The solution organizes feature processing into cardinal groups with internal splits, allowing for both independent feature learning and attention-based recalibration. & The problem is structured as a need to combine multi-branch feature processing with channel-wise attention in a modular and efficient way. The architecture should allow for both parallel feature processing paths and adaptive feature recalibration while maintaining computational efficiency. \\
\textit{Design Rationale:} & \textit{Design Rationale:} \\
Using cardinal groups with internal splits enables parallel processing paths that can learn diverse feature representations while maintaining computational efficiency Implementing a two-level attention mechanism (group-wise and global) allows for modeling both local and global channel interdependencies The hierarchical structure enables the network to capture features at multiple scales while maintaining information flow between different processing paths & Using both cardinal groups and splits within groups allows capturing different types of features at multiple scales/abstraction levels while maintaining computational efficiency through shared operations Applying attention within cardinal groups rather than globally helps maintain specialized feature learning while still allowing adaptive feature combination The hierarchical structure (splits within cardinal groups) provides a good balance between specialized processing and parameter efficiency \\
\textit{Universal Principle:} & \textit{Universal Principle:} \\
Complex systems can be effectively managed by organizing them into hierarchical groups with both local and global optimization mechanisms, allowing for parallel processing while maintaining overall coherence. & Complex information processing can be improved by splitting computation into multiple specialized paths while using context-dependent weighting to adaptively combine their outputs. \\
\textit{Key Mechanism:} & \textit{Key Mechanism:} \\
The network divides input features into C cardinal groups, each containing S split paths. Each split path processes features independently through a sequence of convolutions. Group-wise SE blocks model local channel interdependencies within each cardinal group, while a global SE block captures inter-group dependencies. The final output is produced by aggregating features across all paths using learned attention weights. & Input features are divided into K cardinal groups, each further split into R paths. Each split undergoes separate convolution transformations. Within each cardinal group, splits are combined using attention weights derived from global context. These weights are computed using global pooling followed by two FC layers with softmax to generate split attention factors. Finally, outputs from all cardinal groups are concatenated. \\
\midrule
\multicolumn{2}{l}{\textbf{Case 63}} \\
\textit{Problem Structure:} & \textit{Problem Structure:} \\
The problem is structured as a multi-stage process where an initial model trained on limited labeled data generates pseudo-labels that are then systematically refined and filtered through increasingly strict verification stages, allowing for gradual expansion of usable training data while maintaining quality. & The problem is framed as leveraging a small amount of labeled data to create a baseline model that can generate pseudo-labels for unlabeled audio data. These pseudo-labels, after appropriate filtering, can then be used as additional training data to improve model performance. The structure assumes that imperfect but filtered pseudo-labels can provide useful training signal. \\
\textit{Design Rationale:} & \textit{Design Rationale:} \\
The two-stage verification process (flexible matching followed by strict verification) provides a balance between data quantity and quality, allowing initial broad coverage while ensuring final quality through rigorous filtering. Using attention mechanisms and bidirectional processing enables the model to capture both local and global context, improving transcription accuracy even with limited initial training data. The iterative refinement approach with confidence-based filtering allows the system to gradually improve both the model and pseudo-labels while preventing error propagation. & Using a strong baseline model and language model helps generate higher quality initial pseudo-labels, providing better training signal Filtering mechanisms specific to sequence model errors helps remove common failure modes that could hurt training Balancing pseudo-label quantity vs quality through filtering thresholds allows maximizing useful training signal while minimizing noise \\
\textit{Universal Principle:} & \textit{Universal Principle:} \\
Complex tasks with limited ground truth data can be effectively addressed through iterative bootstrapping processes that gradually expand the reliable data pool while maintaining strict quality controls. & A system can improve its own performance by learning from its filtered predictions on unlabeled data, where appropriate filtering mechanisms help maintain prediction quality while maximizing useful training signal. \\
\textit{Key Mechanism:} & \textit{Key Mechanism:} \\
The system employs an RNN Encoder-Decoder with Gated Hidden Units and attention mechanisms for initial transcription. Generated pseudo-labels undergo two-stage verification using flexible matching and strict verification with a hybrid word/phone graph. Confidence scores and attention-based alignment guide the filtering process. The system iteratively retrains on high-confidence pseudo-labels while maintaining separate data pools based on quality metrics. & 1. Train baseline model on limited labeled data 2. Use baseline model with language model to generate pseudo-labels for unlabeled audio 3. Apply filtering mechanisms:    - Heuristic filters for common sequence model errors (repeating n-grams, incomplete outputs)    - Confidence-based filtering using normalized log likelihood scores 4. Train new model on combined labeled and filtered pseudo-labeled data \\
\midrule


\midrule
\multicolumn{2}{l}{\textbf{Case 66}} \\
\textit{Problem Structure:} & \textit{Problem Structure:} \\
The challenge is structured as a need to increase model depth and expressiveness while maintaining computational efficiency and stable training. The solution approaches this by creating a hierarchical architecture that progressively processes speech at different temporal resolutions, allowing for increased depth without computational explosion. & The problem is framed as a need to increase network depth and non-linearity while managing parameter count and avoiding overfitting. This involves strategically inserting additional processing layers between existing components while maintaining trainability. \\
\textit{Design Rationale:} & \textit{Design Rationale:} \\
Combining small convolutional filters with residual connections allows for very deep networks while maintaining stable gradient flow and controlling parameter count Progressive temporal resolution reduction through pooling enables efficient processing of long sequences while capturing both fine and coarse-grained features Multi-scale feature processing through parallel paths ensures comprehensive capture of speech patterns at different time scales & Adding processing depth through 1x1 convolutions allows the network to learn more complex patterns while keeping parameters manageable through dimensionality reduction Inserting modules between existing layers maintains the basic network structure while providing additional non-linear transformations for better feature learning Using batch normalization in the inserted modules helps manage the training dynamics of the deeper network \\
\textit{Universal Principle:} & \textit{Universal Principle:} \\
Complex hierarchical patterns can be efficiently learned by progressively reducing resolution while maintaining multiple processing scales, allowing depth to be increased without corresponding computational complexity & System expressiveness can be increased by strategically inserting additional processing layers that add depth without excessive parameter growth. \\
\textit{Key Mechanism:} & \textit{Key Mechanism:} \\
The encoder uses stacked 3x3 convolutional layers (50-100 layers) with residual connections, incorporating parallel paths of different kernel sizes (1,3,5,7) at each level. Temporal pooling reduces sequence length by factor n between levels, while auxiliary classifiers at intermediate layers provide additional supervision. The architecture maintains high resolution in early layers for fine acoustic details and reduces resolution progressively for higher-level features. & 1x1 convolution modules (Network-in-Network) are inserted between LSTM layers in the hierarchical subsampling connections. Each module consists of: 1) Frame concatenation 2) Dimension projection 3) Batch normalization 4) ReLU activation. Additional 1x1 convolution modules are also added between LSTM layers to further increase depth. \\
\midrule
\multicolumn{2}{l}{\textbf{Case 67}} \\
\textit{Problem Structure:} & \textit{Problem Structure:} \\
The problem is structured as a two-stage recognition process where acoustic and textual information are processed separately in the first pass and then integrated comprehensively in the second pass, allowing for both efficient initial recognition and thorough refinement. & The problem is framed as needing a second pass that can effectively combine and leverage both acoustic and text information from the first pass to improve recognition accuracy. The structure assumes that having access to both modalities allows for better error correction and sequence refinement by using complementary information. \\
\textit{Design Rationale:} & \textit{Design Rationale:} \\
The dual-attention mechanism allows simultaneous consideration of both acoustic and textual context during refinement, enabling more accurate error correction than methods using only one information source The streaming first-pass design maintains low latency for initial recognition while allowing for more comprehensive processing in the second pass The end-to-end training approach with weighted objectives ensures balanced performance between quick initial recognition and accurate final output & Using both acoustic and text information allows the model to leverage complementary signals - acoustics help with ambiguous pronunciations while text provides linguistic context Bidirectional encoding of first-pass hypotheses enables the model to leverage future context for better refinement of current predictions Separate attention mechanisms for acoustic and text information allow the model to dynamically weight the importance of each information source \\
\textit{Universal Principle:} & \textit{Universal Principle:} \\
Complex sequential decisions can be improved by combining fast initial processing with deliberate refinement using multiple information sources & Complex decision tasks can be improved by multi-pass processing where later stages have access to both original inputs and initial predictions, allowing refinement based on broader context and multiple information sources. \\
\textit{Key Mechanism:} & \textit{Key Mechanism:} \\
The system employs an encoder (E) for acoustic processing, a streaming first-pass decoder (D1) for initial transcription, and a second-pass decoder (D2) with multi-head attention mechanisms attending to both E and D1 outputs. The system is trained end-to-end using Monte Carlo sampling for gradient estimation, optimizing a weighted combination of first-pass and second-pass accuracy. & The model uses a two-pass architecture where: 1) A streaming RNN-T generates initial hypotheses, 2) A deliberation decoder processes these hypotheses using bidirectional encoding and attends to both the encoded hypotheses and original acoustic features through separate attention mechanisms, 3) The attended information is combined to generate refined predictions. The process can be trained end-to-end using a combination of cross-entropy and minimum word error rate losses. \\
\midrule
\multicolumn{2}{l}{\textbf{Case 68}} \\
\textit{Problem Structure:} & \textit{Problem Structure:} \\
The problem is structured as a multi-level challenge requiring simultaneous handling of streaming capabilities, global context integration, and knowledge incorporation from multiple data sources. The solution uses a hierarchical architecture that processes information at different temporal and contextual scales. & The challenge is structured as integrating knowledge from multiple data sources (audio, text, pronunciation) into a single end-to-end model through strategic pretraining and architecture design, while maintaining streaming capability. This requires careful orchestration of how different types of knowledge are injected into different parts of the model. \\
\textit{Design Rationale:} & \textit{Design Rationale:} \\
The hierarchical structure allows for efficient streaming processing at lower levels while incorporating broader context and additional knowledge sources at higher levels Using chunk-based processing with local and global attention mechanisms balances real-time requirements with the need for contextual information The multi-task learning approach enables effective incorporation of knowledge from different data sources while maintaining model coherence & Pretraining components on specialized data sources allows the model to benefit from larger amounts of data than would be available for end-to-end training alone The hierarchical pretraining structure helps inject domain knowledge about the natural decomposition of the problem (acoustics -> phonemes -> graphemes/wordpieces) Maintaining streaming capability by using an architecture (RNN-T) that can process inputs incrementally rather than requiring the full sequence \\
\textit{Universal Principle:} & \textit{Universal Principle:} \\
Complex systems can be effectively managed by processing information at multiple scales simultaneously, with different levels handling different aspects of the problem while maintaining communication between levels. & Complex transformation systems can be improved by independently pretraining specialized components on relevant subtasks before integration, allowing each component to develop robust foundations in its specific role. \\
\textit{Key Mechanism:} & \textit{Key Mechanism:} \\
The model uses a chunk-based RNN Transducer at the lower level for streaming acoustic processing, combined with a higher-level network that incorporates global context through attention mechanisms over cached representations. The decoder uses a shared subword vocabulary created using WPM. Multi-task learning objectives combine acoustic modeling, pronunciation prediction, and language modeling tasks. The model processes input in fixed-size chunks, maintaining a cache of hidden states for global context access. & The system consists of an encoder network pretrained with CTC loss on acoustic data (optionally with hierarchical phoneme/grapheme losses), and a decoder network pretrained as a language model on text data. These components are then combined and jointly fine-tuned using the RNN-T loss to create an integrated streaming speech recognition system. \\
\midrule
\multicolumn{2}{l}{\textbf{Case 69}} \\
\textit{Problem Structure:} & \textit{Problem Structure:} \\
The problem is structured as an end-to-end optimization challenge where phase reconstruction needs to be incorporated directly into the training process, rather than being treated as a separate post-processing step. This allows the model to learn magnitude estimates that are inherently optimized for phase reconstruction. & The problem is structured as jointly optimizing magnitude estimation and phase reconstruction by incorporating iterative phase reconstruction algorithms directly into the training process. This allows the magnitude estimation to be aware of and optimize for subsequent phase reconstruction steps. \\
\textit{Design Rationale:} & \textit{Design Rationale:} \\
Implementing phase reconstruction as differentiable operations allows gradients to flow through the entire pipeline, enabling true end-to-end optimization Using a two-stage architecture with shared initial layers balances computational efficiency with the need for specialized processing of magnitude and phase information Incorporating a confidence-based weighting mechanism helps focus the optimization on regions where phase reconstruction is most critical & Incorporating phase reconstruction into training allows the magnitude estimation to anticipate and compensate for the behavior of phase reconstruction, leading to better overall results than treating them as separate steps Using curriculum learning (progressively increasing the number of unfolded iterations) helps manage the complexity of training through multiple iterations Computing loss on the final time-domain signal provides a more direct optimization target that accounts for both magnitude and phase accuracy \\
\textit{Universal Principle:} & \textit{Universal Principle:} \\
Complex multi-step processes can be optimized holistically by making each step differentiable and allowing feedback from later stages to influence earlier stages during training & End-to-end optimization of multi-stage processes can achieve better results than optimizing each stage independently, particularly when later stages can influence how earlier stages should behave. \\
\textit{Key Mechanism:} & \textit{Key Mechanism:} \\
The solution implements a differentiable version of iterative phase reconstruction algorithms as neural network layers. The pipeline consists of: 1) magnitude estimation network, 2) differentiable phase reconstruction module with confidence-based updating, 3) time-domain loss computation. During training, gradients flow backward through all components, allowing the magnitude estimation to be optimized while considering its impact on phase reconstruction. & The iterative phase reconstruction algorithm (MISI) is unfolded as deterministic layers in the neural network, allowing backpropagation through the entire process. This enables the magnitude estimation network to learn to produce estimates that are well-suited for subsequent phase reconstruction. The loss is computed on the final time-domain signal after phase reconstruction rather than on intermediate representations. \\
\midrule
\multicolumn{2}{l}{\textbf{Case 70}} \\
\textit{Problem Structure:} & \textit{Problem Structure:} \\
The problem is structured as a continuous information accumulation and boundary detection task, where acoustic features need to be processed sequentially while maintaining online capability and accurate boundary positioning. The solution combines monotonic attention for sequential processing with neural accumulation mechanisms for boundary detection. & The problem is framed as creating a soft, monotonic alignment mechanism that incrementally accumulates acoustic information until sufficient evidence is gathered to emit an output token, similar to how biological neurons integrate inputs until reaching a firing threshold. \\
\textit{Design Rationale:} & \textit{Design Rationale:} \\
Combining monotonic attention with soft attention over small chunks ensures both online processing capability and effective feature integration Using an adaptive neural accumulator with learned firing thresholds allows the system to dynamically adjust to different speech patterns Implementing a differentiable boundary detection mechanism enables end-to-end training while maintaining biological plausibility & The continuous, differentiable nature of the mechanism allows end-to-end training while maintaining the ability to make discrete decisions, bridging the gap between continuous input processing and discrete output generation The forward-only, monotonic processing enables online/streaming applications while maintaining efficiency by avoiding unnecessary computations on irrelevant inputs The explicit boundary detection and information splitting at transition points ensures proper utilization of all input information and provides interpretable alignment points \\
\textit{Universal Principle:} & \textit{Universal Principle:} \\
Complex sequential processing tasks can be effectively handled through a combination of strict forward progression (monotonicity) and adaptive local integration, mimicking biological neural systems' ability to accumulate information until reaching meaningful decision points. & Continuous accumulation of information with threshold-based discretization provides an efficient way to align and transform sequences of different granularities. \\
\textit{Key Mechanism:} & \textit{Key Mechanism:} \\
The system processes input using monotonic attention while accumulating acoustic information through a neural accumulator. A 1D CNN with sigmoidal unit computes halting probabilities that are accumulated until reaching an adaptive threshold. When fired, the accumulated information is processed using soft attention over the local chunk for recognition. The entire process remains differentiable through the use of expected values during training. & At each encoder step, CIF receives a vector representation and a corresponding weight that scales the information content. It forwardly accumulates weights and integrates vector information until reaching a threshold $\beta$ (typically 1.0), signaling an acoustic boundary. When a boundary is detected, the acoustic information is divided into two parts - one completing the current token integration and one starting the next - before firing the integrated embedding to the decoder. \\
\midrule
\multicolumn{2}{l}{\textbf{Case 71}} \\
\textit{Problem Structure:} & \textit{Problem Structure:} \\
The problem is structured as a dual challenge of maintaining low latency while preserving model quality in streaming speech recognition. The solution combines triggered processing for efficiency with local attention windows for maintaining context quality. & The problem is framed as a need to process input sequences in fixed-length chunks while maintaining the powerful representational capabilities of transformers. This requires restructuring the attention mechanism to work with limited context and developing a synchronized encoding-decoding process that can generate outputs incrementally. \\
\textit{Design Rationale:} & \textit{Design Rationale:} \\
Combining CTC triggering with local attention windows provides an optimal balance between processing efficiency and context preservation Using a Gaussian-based local attention mechanism allows for flexible yet controlled attention span within chunks The hybrid approach reduces computational overhead while maintaining the ability to capture important speech patterns & Restricting attention to left contexts enables real-time processing by eliminating dependency on future inputs while still allowing modeling of long-term dependencies through accumulated context Overlapping chunks maintain smooth information flow between windows, preventing loss of context at chunk boundaries Forward-backward algorithm optimizes all possible alignments rather than just the best path, leading to more robust output generation \\
\textit{Universal Principle:} & \textit{Universal Principle:} \\
Complex sequential processing tasks can be optimized by combining efficient trigger mechanisms for identifying key processing points with local context windows for maintaining quality. & Complex sequence transformation systems can be made to work incrementally by carefully managing information flow between fixed-size processing windows while maintaining sufficient context for coherent output generation. \\
\textit{Key Mechanism:} & \textit{Key Mechanism:} \\
The system uses a CTC network to trigger processing at likely phoneme boundaries, then applies a local attention window with Gaussian distribution around these points. The attention window spans the current chunk plus a small look-ahead buffer. The context vector is computed as a weighted combination of the local attention scores and the Gaussian prior. & The model processes input in fixed-length chunks with overlap between adjacent chunks. The encoder restricts self-attention to only left contexts, eliminating dependency on future information. The decoder generates outputs chunk-by-chunk using a forward-backward algorithm to optimize all possible alignment paths between input chunks and output symbols. The overlap between chunks (typically 20\% of chunk length) helps maintain smooth information transition. \\
\midrule
\multicolumn{2}{l}{\textbf{Case 72}} \\
\textit{Problem Structure:} & \textit{Problem Structure:} \\
The problem is structured as a multi-objective optimization challenge where different aspects of facial animation (identity, expressions, lip sync) need to be generated and controlled independently before being combined coherently. This allows for specialized handling of each component while maintaining overall consistency. & The problem is framed as needing two complementary quality assessment mechanisms - one for evaluating individual frame realism and identity preservation, and another for assessing temporal coherence and audio-visual synchronization across sequences. This allows addressing both spatial and temporal aspects of video generation simultaneously. \\
\textit{Design Rationale:} & \textit{Design Rationale:} \\
Separating the generation process into multiple streams allows each component to be optimized independently while maintaining coordination through learned masks Using different types of convolutions (2D for identity, 3D for dynamics) optimizes the architecture for each specific task Incorporating audio features directly into the dynamic streams ensures tight synchronization while allowing for natural variation & Using two discriminators allows specialized focus on both spatial and temporal quality aspects without requiring either discriminator to evaluate all quality criteria The RNN-based sequence discriminator enables evaluation of variable-length sequences while maintaining causal generation capability needed for real-time applications The frame discriminator helps maintain consistent identity and visual quality while allowing natural variations in expression \\
\textit{Universal Principle:} & \textit{Universal Principle:} \\
Complex systems can be decomposed into specialized components that operate independently but are integrated through learned coordination mechanisms to produce coherent output & Complex generative tasks can be better controlled by using multiple specialized evaluation mechanisms that each focus on different quality aspects of the output. \\
\textit{Key Mechanism:} & \textit{Key Mechanism:} \\
The system uses three parallel streams: (1) an identity stream with 2D convolutions generating static facial features, (2) an expression stream with 3D spatio-temporal convolutions for dynamic expressions, and (3) a lip sync stream processing audio features for mouth movements. These are combined using learned spatio-temporal masks that determine each stream's contribution per region and timestep. & The system uses two discriminators: 1) A Frame Discriminator that ensures photorealism and identity preservation in individual frames through adversarial training on image pairs. 2) A Sequence Discriminator that evaluates temporal coherence and audio-visual synchronization across frame sequences using RNNs. The generator is trained to simultaneously satisfy both discriminators while maintaining reconstruction quality. \\
\midrule
\multicolumn{2}{l}{\textbf{Case 73}} \\
\textit{Problem Structure:} & \textit{Problem Structure:} \\
The problem is structured as two independent representation learning tasks (text and speech) followed by an alignment phase, allowing each component to be optimized separately with unpaired data before being connected with limited paired data. & The problem is structured as enhancing the text encoder's representational capabilities by incorporating external textual knowledge from large unpaired text corpora, while maintaining the ability to learn alignments with acoustic features. \\
\textit{Design Rationale:} & \textit{Design Rationale:} \\
Independent pre-training of text and speech encoders allows leveraging large amounts of unpaired data to learn rich representations before attempting alignment Using Global Style Tokens for both text and speech enables capturing stylistic variations that can be aligned later, improving expressiveness The soft attention mechanism with Gaussian windows provides a flexible yet constrained way to align representations during fine-tuning & Conditioning on pre-trained word vectors allows the model to leverage rich linguistic knowledge without requiring additional paired data collection The flexibility in conditioning location (input vs top) and method (concatenation vs attention) allows for different trade-offs between information flow and parameter efficiency \\
\textit{Universal Principle:} & \textit{Universal Principle:} \\
Complex learning tasks can be decomposed into independent representation learning phases followed by alignment, allowing efficient use of abundant unpaired data while minimizing the need for paired data. & Domain-specific knowledge learned from large unpaired datasets can be effectively transferred to enhance representation learning in systems with limited paired training data. \\
\textit{Key Mechanism:} & \textit{Key Mechanism:} \\
The system uses LSTM encoders with Global Style Tokens for both text and speech domains, pre-trained independently on unpaired data. These are then aligned using a soft attention mechanism with Gaussian windows during fine-tuning with limited paired data. A Mixture Density Network generates the final speech output, conditioned on both aligned representations and style tokens. & External word vectors are incorporated into the encoder through either concatenation at word boundaries or attention-based conditioning. The vectors can be added either at the encoder input (phoneme embedding level) or encoder top (final representation level). The model then fine-tunes these representations using the limited paired data to learn the text-to-speech alignment. \\
\midrule


\multicolumn{2}{l}{\textbf{Case 76}} \\
\textit{Problem Structure:} & \textit{Problem Structure:} \\
The problem is approached by treating code layout optimization as a trace-based organization challenge that simultaneously considers branch prediction accuracy and cache performance. It recognizes that branch mispredictions often stem from poor code organization that fails to align with actual execution patterns. & The challenge is structured around optimizing code layout to influence branch behavior and prediction patterns in a way that reduces negative interference while maintaining good branch distribution characteristics. This involves analyzing and modifying both static and dynamic aspects of branch prediction. \\
\textit{Design Rationale:} & \textit{Design Rationale:} \\
Combining profile information with trace construction allows for data-driven optimization that reflects actual program behavior rather than static assumptions Using a software-level trace cache approach provides the benefits of trace-based execution without requiring specialized hardware support Integrating cache-aware mapping ensures that optimizations for branch prediction don't come at the cost of cache performance & Making branches tend toward not-taken behavior reduces negative interference in prediction tables since conflicts between branches are more likely to reinforce rather than contradict each other Optimizing code layout improves instruction cache performance and fetch bandwidth, which can outweigh potential losses in prediction accuracy for some predictors The approach balances multiple performance factors (prediction accuracy, cache performance, fetch bandwidth) rather than optimizing for prediction accuracy alone \\
\textit{Universal Principle:} & \textit{Universal Principle:} \\
Optimizing the physical organization of frequently accessed elements based on their usage patterns can significantly improve system performance by reducing access conflicts and prediction errors. & System element organization can be optimized to improve predictability by reducing negative interference patterns, though this may create new trade-offs that need to be balanced against overall system goals. \\
\textit{Key Mechanism:} & \textit{Key Mechanism:} \\
The solution uses profile data to identify frequent execution paths and constructs traces by grouping commonly executed sequences. These traces are then mapped to memory locations in a cache-conscious way, with selective code replication for critical paths. The layout process specifically considers branch predictor table aliasing patterns to minimize harmful interference. & The code layout optimization works by reordering basic blocks to make branches tend toward not-taken behavior. This reduces negative interference in branch prediction tables since branches mapping to the same entry are more likely to push the prediction counter in the same direction. However, it also affects the distribution of branch history register values, which can impact some prediction schemes. \\
\midrule
\multicolumn{2}{l}{\textbf{Case 77}} \\
\textit{Problem Structure:} & \textit{Problem Structure:} \\
The problem is structured as a multi-stream processing challenge where different modalities must be processed independently while maintaining temporal synchronization and allowing for both cross-modal interactions and preservation of modality-specific features. & The problem is structured as a multi-modal fusion challenge where features from different modalities need to be combined effectively while preserving their complementary discriminative power. The structure assumes that deception manifests through multiple channels simultaneously and that their joint analysis provides better detection accuracy than individual modalities. \\
\textit{Design Rationale:} & \textit{Design Rationale:} \\
The parallel processing streams with modality-specific architectures ensure optimal feature extraction for each data type while maintaining temporal alignment through a central coordination system The dual-path fusion approach allows for both element-wise interactions between compatible modalities and preservation of unique modal characteristics through concatenation The timestamp-based synchronization system enables precise temporal alignment of features across modalities with different sampling rates & Using specialized feature extractors for each modality allows capturing modality-specific patterns effectively while keeping the overall architecture simple The fusion strategy (concatenation vs Hadamard+concatenation) allows exploring different ways of combining modalities while controlling model complexity \\
\textit{Universal Principle:} & \textit{Universal Principle:} \\
Complex multi-modal systems can be effectively integrated by maintaining independent processing streams while using a centralized coordination mechanism to ensure synchronization and selective interaction between components. & Complex behavioral patterns can be better detected by analyzing multiple complementary channels of information simultaneously rather than looking at individual channels in isolation. \\
\textit{Key Mechanism:} & \textit{Key Mechanism:} \\
The system implements parallel processing streams for each modality (video, audio, text, micro-expressions) using specialized architectures (3D CNNs for video, 1D CNNs for audio, etc.). A central timestamp-based ring buffer coordinates data flow and synchronization. Each modality maintains dual channels (static and trainable) to preserve general knowledge while allowing task-specific adaptation. Cross-modal fusion occurs through both element-wise interaction layers for compatible modalities and concatenation layers for preserving unique modal information. & The system extracts features from each modality using specialized neural networks (3D-CNN for video, CNN for text, openSMILE for audio) and combines them using either concatenation or Hadamard product followed by concatenation with micro-expression features. The fused features are then passed through a simple MLP classifier. \\
\midrule

\midrule
\multicolumn{2}{l}{\textbf{Case 79}} \\
\textit{Problem Structure:} & \textit{Problem Structure:} \\
The problem is structured as a hierarchical fusion challenge where information from different sensor modalities is combined at multiple levels, starting from generating standardized representations and progressively refining detections through geometric and semantic consistency checks. & The problem is framed as a late fusion approach that leverages geometric and semantic consistencies between pre-NMS detection candidates from both modalities. Rather than forcing early feature alignment or complex deep fusion, it focuses on learning optimal combination rules at the detection level. \\
\textit{Design Rationale:} & \textit{Design Rationale:} \\
Using multi-view representations (bird's eye view, front view, and image view) allows the system to capture complementary information from different perspectives while maintaining a consistent data structure Implementing a hierarchical fusion approach with both early and late fusion components enables the system to leverage both low-level feature combinations and high-level semantic information Incorporating explicit geometric and semantic consistency checks ensures that the final detections are reliable and well-aligned across modalities & Late fusion allows use of pre-trained single-modality detectors without modification or retraining, making the system modular and practical to deploy with different detector combinations Operating on pre-NMS candidates with low thresholds maximizes recall and allows the fusion network to learn optimal suppression patterns using cross-modal information Sparse tensor processing maintains efficiency while capturing all relevant detection candidate relationships \\
\textit{Universal Principle:} & \textit{Universal Principle:} \\
Complex multi-modal information can be effectively combined by creating standardized intermediate representations and implementing progressive refinement through multiple stages of fusion and consistency checking. & Late-stage fusion of preliminary decisions/candidates using learned consistency patterns can effectively combine complementary information sources while avoiding complex alignment and architecture issues. \\
\textit{Key Mechanism:} & \textit{Key Mechanism:} \\
The solution first generates detection candidates separately from each modality, projects them into multiple standardized views, extracts multi-scale features using an encoder-decoder architecture with skip connections, and then performs progressive fusion through geometric alignment, semantic consistency checking, and final refinement. The fusion process is guided by confidence scores and cross-modality consistency metrics. & Creates a sparse tensor encoding geometric (IoU) and semantic consistencies between all pairs of 2D and 3D detection candidates, processes non-empty elements using 1x1 convolutions to learn optimal fusion patterns, and generates final 3D detection scores through maxpooling. Key components: 1. Input tensor $T_{i,j} = \{IoU_{i,j}, s_{2D}[i], s_{3D}[j], d[j]\}$ for each 2D-3D candidate pair 2. Four 1x1 conv layers process non-empty elements  3. Maxpooling generates final scores while maintaining sparsity \\
\midrule
\multicolumn{2}{l}{\textbf{Case 80}} \\
\textit{Problem Structure:} & \textit{Problem Structure:} \\
The problem is structured as a hierarchical graph processing task where videos are represented as spatial-temporal graphs with both local (frame-level) and global (video-level) relationships. The solution addresses the challenge through a multi-scale approach that combines efficient graph processing with residual learning. & The problem is structured as a graph representation challenge where nodes represent object regions and edges capture both spatial relationships within frames and temporal relationships across adjacent frames. The structure assumes locality in relationships and organizes frames as sub-graphs connected through temporal edges. \\
\textit{Design Rationale:} & \textit{Design Rationale:} \\
The hierarchical structure allows efficient processing of both spatial and temporal relationships while maintaining computational feasibility Residual learning enables the model to focus on learning changes in relationships rather than absolute states, making it particularly effective for temporal modeling The bottleneck architecture reduces computational complexity while preserving essential relationship information & Locality constraints reduce computational complexity while preserving important local relationships Hierarchical sub-graph organization allows capturing both fine-grained object relationships and higher-level temporal patterns Learned refinement enables discovery of task-relevant relationships while removing noise \\
\textit{Universal Principle:} & \textit{Universal Principle:} \\
Complex temporal-spatial relationships can be efficiently modeled by focusing on incremental changes and using hierarchical processing structures that separate local and global patterns. & Complex sequential relationships can be effectively modeled through hierarchical graph structures with learned refinements that balance local and global dependencies. \\
\textit{Key Mechanism:} & \textit{Key Mechanism:} \\
The solution implements a three-level hierarchical transformer architecture: 1) Within-frame spatial attention using bottleneck compression (1x1 conv → graph attention → 1x1 conv), 2) Cross-frame temporal attention with residual connections G(t) = F(G(t-1)) + G(t-1), and 3) Global video-level attention for long-range dependencies. Each level uses multi-head attention with learned weight vectors for computing relationship strengths. & Constructs an initial graph with nodes representing object regions and edges between spatially/temporally adjacent regions. Only allows connections between objects in adjacent frames. Refines graph topology through learned similarity metrics and pruning. Graph structure follows principles of locality and hierarchical organization through sub-graphs. \\
\midrule
\multicolumn{2}{l}{\textbf{Case 81}} \\
\textit{Problem Structure:} & \textit{Problem Structure:} \\
The problem is reframed as transforming complex indirect branch prediction into a series of simpler binary decisions that can leverage existing conditional branch prediction hardware. This approach eliminates the need for large dedicated hardware structures while maintaining prediction accuracy. & The problem is framed as one of mapping an N-way prediction problem (indirect branch target prediction) to a series of simpler binary predictions (conditional branch predictions) that can leverage existing hardware. This structure assumes that complex N-way decisions can be decomposed into a sequence of simpler binary choices while maintaining accuracy. \\
\textit{Design Rationale:} & \textit{Design Rationale:} \\
Using decision trees allows for efficient binary choices that can be handled by existing conditional branch predictors, reducing hardware complexity Profile-guided optimization ensures the most common targets are checked first, improving average-case performance Dynamic adaptation capabilities allow the system to maintain accuracy as program behavior changes over time & Using existing conditional branch prediction hardware avoids the need for separate complex prediction structures for indirect branches, significantly reducing hardware cost and complexity compared to specialized predictors The iterative prediction process with virtual branches allows the system to adapt dynamically to different numbers of targets per indirect branch without needing static analysis or profiling Deriving virtual PCs through hashing helps distribute predictions across the prediction tables while maintaining correlation between iterations for the same indirect branch \\
\textit{Universal Principle:} & \textit{Universal Principle:} \\
Complex multi-choice decisions can be decomposed into a series of binary choices, allowing simpler existing mechanisms to handle complex scenarios efficiently. & Complex multi-way decisions can be effectively transformed into sequences of simpler binary decisions by creating virtual intermediate states that exist only in the prediction/control hardware. \\
\textit{Key Mechanism:} & \textit{Key Mechanism:} \\
Creates a profile-guided decision tree where each node represents a potential target. The indirect branch is transformed into a series of conditional branches that traverse this tree. Each decision point uses branch history context for prediction. The tree structure is dynamically adjusted based on runtime behavior, with most frequent targets placed near the root. & Each indirect branch is treated as a sequence of virtual conditional branches for prediction purposes. Each virtual branch has a unique virtual PC (VPCA) and virtual global history register (VGHR) derived from the original PC and GHR. The predictor iteratively checks each virtual branch using VPCA/VGHR values, continuing until either a taken prediction is found (providing the target) or a maximum number of iterations is reached. The VPCA for iteration N is computed as PC  HASHVAL[N] where HASHVAL is a table of randomized values. \\
\midrule



\multicolumn{2}{l}{\textbf{Case 85}} \\
\textit{Problem Structure:} & \textit{Problem Structure:} \\
The problem is structured as a dual challenge of capturing both content and relational information in author disambiguation without complex feature engineering. The solution uses an adversarial framework where a generator explores the network space while a discriminator learns to distinguish authentic authorship patterns. & The problem is structured as an adversarial learning framework where a discriminative module learns to distinguish paper pairs from the same author while a generative module selects potentially homogeneous papers from a heterogeneous information network. The two modules compete and cooperate to improve each other's performance. \\
\textit{Design Rationale:} & \textit{Design Rationale:} \\
The adversarial approach naturally integrates both content and structural information without explicit feature engineering, as the generator and discriminator learn to focus on relevant patterns through competition Using policy gradient methods in the generator allows for effective exploration of discrete author-paper associations while learning from discriminator feedback The framework can capture high-order connections naturally through the iterative adversarial process, as the generator learns to explore complex paths in the heterogeneous network & Using adversarial training allows the model to leverage both supervised pattern recognition (via discriminator) and unsupervised structure discovery (via generator) in a unified framework The random walk based generator efficiently explores high-order connections in the network while being guided by the discriminator's feedback, avoiding the need for manual feature engineering Self-training of the discriminator reduces dependency on labeled data while maintaining high precision through iterative refinement \\
\textit{Universal Principle:} & \textit{Universal Principle:} \\
Complex pattern recognition tasks can be solved through adversarial learning where competing modules simultaneously explore and validate potential solutions, eliminating the need for manual feature engineering. & Competitive interaction between local pattern recognition and global structure exploration can help discover complex relationships without extensive supervision or feature engineering. \\
\textit{Key Mechanism:} & \textit{Key Mechanism:} \\
The framework consists of a generator G that uses policy gradient methods to select author-paper pairs from a heterogeneous information network, and a discriminator D that learns to distinguish true pairs. The objective is formulated as: $$\min_G \max_D V(D,G) = \mathbb{E}[\log D(x)] + \mathbb{E}[\log(1-D(G(z)))]$$ where x represents true author-paper pairs and G(z) represents generated pairs. The generator's policy is updated using REINFORCE algorithm with the discriminator's output as reward. & The framework consists of two key modules: (1) A discriminative module that uses a neural network to distinguish whether paper pairs are from the same author based on content and relation embeddings. (2) A generative module that uses random walks on a heterogeneous information network to select potentially homogeneous papers. The modules are trained adversarially - the discriminator provides rewards to guide the generator's paper selection, while the generator creates challenging negative samples to improve the discriminator's ability to capture high-order connections. \\
\midrule
\multicolumn{2}{l}{\textbf{Case 86}} \\
\textit{Problem Structure:} & \textit{Problem Structure:} \\
The problem is structured as a multi-level representation challenge where long-term preferences, short-term preferences, and query intent need to be dynamically weighted and combined. The solution uses hierarchical embeddings with attention mechanisms to selectively focus on relevant aspects of each preference type based on the current query context. & The problem is structured as a need to model and combine three key information sources: 1) The current search query intent, 2) The user's relatively stable long-term preferences that evolve slowly over time, and 3) The user's dynamic short-term preferences inferred from recent purchases. Each of these components needs to be weighted appropriately based on their relevance to the current search context. \\
\textit{Design Rationale:} & \textit{Design Rationale:} \\
Using separate embedding spaces for long-term and short-term preferences allows the model to capture different temporal aspects of user behavior without interference The attention mechanism with query as the key enables dynamic weighting of preferences based on their relevance to the current search intent Non-linear transformations of embeddings help capture complex relationships between preferences and queries & The dual attention mechanisms allow the model to selectively focus on relevant aspects of both long-term and short-term preferences, preventing irrelevant historical information from affecting the current prediction The gradual updating of long-term preferences balances stability with adaptability, allowing the model to maintain consistent user profiles while slowly incorporating new behavioral patterns Using GRU for short-term preference modeling captures sequential patterns in recent user actions while maintaining computational efficiency \\
\textit{Universal Principle:} & \textit{Universal Principle:} \\
Complex temporal behaviors can be effectively modeled by separating them into different timescales and using context-dependent attention mechanisms to dynamically combine them. & When analyzing temporal behavioral patterns, both stable long-term characteristics and dynamic short-term trends should be considered, with their relative importance determined by the current context \\
\textit{Key Mechanism:} & \textit{Key Mechanism:} \\
The solution creates three embedding spaces: long-term preferences (ulong), short-term preferences (ushort), and query (q). These are transformed using non-linear projections: u'long = tanh(Wlong·ulong + blong), u'short = tanh(Wshort·ushort + bshort). The final representation is computed as: r = attention(q, [u'long, u'short]) where attention weights are calculated using query relevance scores. & The model consists of three main components: 1) Attentive Short-Term Preference Modeling using GRU and attention mechanisms to capture recent purchase patterns, 2) Attentive Long-Term Preference Modeling that gradually updates based on purchase history and uses attention to identify relevant aspects, 3) Query Representation Integration that fuses the weighted preferences with the current query through deep neural networks. The final relevance score is computed as the cosine similarity between the integrated query representation and product embeddings. \\
\midrule
\multicolumn{2}{l}{\textbf{Case 87}} \\
\textit{Problem Structure:} & \textit{Problem Structure:} \\
The problem is structured as a dual-objective challenge where content-based and relation-based signals need to be balanced through an adversarial framework. The generator explores network topology while the discriminator learns fine-grained content distinctions, creating a self-improving system that leverages both approaches. & The problem is framed as an adversarial learning task where a discriminative module learns to distinguish papers by the same author while a generative module explores the heterogeneous information network to find potentially homogeneous papers. This allows unsupervised learning of both local content similarities and global network relationships. \\
\textit{Design Rationale:} & \textit{Design Rationale:} \\
The adversarial framework naturally balances exploration of network connections (generator) with precise disambiguation (discriminator) without requiring manual feature engineering Using policy gradient training for the generator allows it to learn effective exploration strategies in the complex heterogeneous network space The iterative feedback loop between generator and discriminator enables continuous improvement of both network topology understanding and content-based discrimination & The adversarial framework allows integration of content and relation information without requiring extensive labeled data, as the generator can explore the network to find similar papers while the discriminator provides supervision signals Using a heterogeneous information network representation eliminates the need for complex feature engineering while capturing rich relationship information between papers through various types of connections The random walk based generator efficiently explores high-order connections in the network while being guided by the discriminator's feedback about paper similarity \\
\textit{Universal Principle:} & \textit{Universal Principle:} \\
Complex pattern recognition tasks can be solved through adversarial learning where complementary approaches are pitted against each other to achieve continuous mutual improvement. & Complex entity disambiguation can be achieved through adversarial learning that combines local feature similarity and global network connectivity without requiring extensive supervision. \\
\textit{Key Mechanism:} & \textit{Key Mechanism:} \\
The generator network explores the heterogeneous author-paper network using policy gradient methods to propose potentially homogeneous paper sets, while the discriminator uses both content features and network topology to distinguish true same-author papers. The two networks are trained iteratively, with the generator maximizing the probability of fooling the discriminator and the discriminator minimizing its error rate. & The framework has two main components: 1) A discriminative module that learns to distinguish if two papers are by the same author based on content and relation embeddings, trained through self-training on pseudo-positive samples. 2) A generative module that explores the heterogeneous network through random walks to find potentially homogeneous papers, guided by rewards from the discriminator. The two modules are trained adversarially, with the generator trying to find papers that can fool the discriminator. \\
\midrule
\end{longtable}

\end{document}